\documentclass[bst/sn-basic]{sn-jnl}

\usepackage[utf8]{inputenc}
\usepackage{graphicx}
\usepackage[T1]{fontenc}
\usepackage{makecell}

\newcommand{\etal}{{\em et al.}}       
\newcommand{\eg}{{\em e.g.}}           
\newcommand{\ie}{{\em i.e.}}           

\newcommand{\ourmodel}{{LigPose~}}
\newcommand{\ourmodelnoblank}{{LigPose}}
\newcommand{\ourmodelscore}{{$\text{\ourmodelnoblank}^{\text{score}}$~}}
\newcommand{\ourmodelscorenoblank}{{$\text{\ourmodelnoblank}^{\text{score}}$}}
\newcommand{\ourmodelaffinity}{{$\text{\ourmodelnoblank}^{\text{affinity}}$~}}
\newcommand{\ourmodelaffinitynoblank}{{$\text{\ourmodelnoblank}^{\text{affinity}}$}}
\newcommand{\ourmodellight}{{$\text{\ourmodelnoblank}^{\text{light}}$}}
\newcommand{\mpronoblank}{{$\text{M}^{\text{pro}}$}}
\newcommand{\mpro}{{$\text{M}^{\text{pro}}$~}}

\jyear{2022}%

\theoremstyle{thmstyleone}%
\theoremstyle{thmstyletwo}%

\theoremstyle{thmstylethree}%

\usepackage{pifont}

\begin{document}
\title[LigPose]{One-step Structure Prediction and Screening for Protein-Ligand Complexes using Multi-Task Geometric Deep Learning}


\author[1,2]{\fnm{Kelei} \sur{He}}\email{hkl@nju.edu.cn}
\equalcont{These authors contributed equally to this work.}

\author[2,3]{\fnm{Tiejun} \sur{Dong}}\email{tiejundong@smail.nju.edu.cn}
\equalcont{These authors contributed equally to this work.}

\author*[1,3]{\fnm{Jinhui} \sur{Wu}}
\email{jfzhang@nju.edu.cn}

\author*[1,2,3]{\fnm{Junfeng} \sur{Zhang}}\email{wuj@nju.edu.cn}

\affil[1]{\orgdiv{State Key Laboratory of Pharmaceutical Biotechnology}, \orgname{Medical School of Nanjing University}, \orgaddress{\city{Nanjing}, \country{P. R. China}}}

\affil[2]{\orgdiv{National Institute of Healthcare Data Science}, \orgname{Nanjing University}, \orgaddress{\city{Nanjing}, \country{P. R. China}}}

\affil[3]{\orgdiv{School of Life Sciences}, \orgname{Nanjing University}, \orgaddress{\city{Nanjing}, \country{P. R. China}}}


\abstract{
Understanding the structure of the protein-ligand complex is crucial to drug development. Existing virtual structure measurement and screening methods are dominated by docking and its derived methods combined with deep learning. However, the sampling and scoring methodology have largely restricted the accuracy and efficiency. Here, we show that these two fundamental tasks can be accurately tackled with a single model, namely LigPose, based on multi-task geometric deep learning. By representing the ligand and the protein pair as a graph, LigPose directly optimizes the three-dimensional structure of the complex, with the learning of binding strength and atomic interactions as auxiliary tasks, enabling its one-step prediction ability without docking tools. Extensive experiments show LigPose achieved state-of-the-art performance on major tasks in drug research. Its considerable improvements indicate a promising paradigm of AI-based pipeline for drug development.
}

\maketitle

\section{Introduction}\label{sec1}



Small organic molecule (SOM) plays an important role in clinical treatment, accounting for about $72\%$ FDA-approved drugs in $2018\sim2022$\cite{mullard20232022}.
Its efficacy is achieved by binding to the target (usually a protein) as a ligand, to produce a protein-ligand complex. 
Understanding the structural details of the protein-ligand complexes at the atomic level reveals the bioactivities of the proteins and ligands, thus guiding the structure-based drug development, \eg, drug screening\cite{lyu2019ultra,zheng2018computational,kitchen2004docking} and lead optimization\cite{kitchen2004docking,zheng2018computational,wang2019structural}. Conventional methods use experimental measurements (\eg, X-ray diffraction\cite{zheng2015x} and cryo-electron microscopy\cite{renaud2018cryo}) to analyze novel protein-ligand complexes, however, are time and resource-expensive.

To alleviate this problem, virtual measurement by molecular docking\cite{pinzi2019molecular,lyu2019ultra,meng2011molecular,kitchen2004docking} has been widely adopted in the past decades to predict the native-like ligand-binding conformations with the respective protein-binding sites. Typically, a docking tool first samples a set of binding conformations (namely poses), and then ranks them using a scoring function to select a top-scored pose\cite{meng2011molecular} (Fig. \ref{fig:framework}\textbf{b}). As reported, popular docking tools generate native-like poses with accuracies from approximately $40\%$ to $60\%$ in terms of success rate\cite{wang2016comprehensive}, which were far from satisfactory. Since deep learning has impacted the field of drug development, many researchers attempted to build hybrid methods by regarding it as a more expressive scoring function to rank the poses sampled by conventional docking tools\cite{IGN,mendez2021geometric,lim2019predicting,DeepBSP,morrone2020combining,ragoza2017protein}. Nevertheless, the performance of these hybrid methods has a constrained upper bound, as limited by the sampling time and space of docking tools\cite{morrone2020combining}.

Recently, several deep learning methods, \ie, AlphaFold\cite{AF2}, RoseTTAFold\cite{RoseTTAFold}, and ESMFold\cite{ESMFold}, have shown great capacity to predict protein structures that outperformed previous methods by a large margin. These methods provided a novel computational approach to generate protein structures directly from their chemical sequences, rethinking the structure prediction paradigm in a data-driven perspective without following the conventional force-field assumption\cite{baek2022deep}. 
Recent progresses\cite{baker_complex,prot_pep,evans2021protein,bryant2022improved} demonstrated that using these methods as the protein/peptide structure estimator can predict precise structures for amino acid-based complexes such as the protein-protein and protein-peptide complexes, showing the generalizability of deep learning methods to the downstream tasks. However, these methods are not specifically designed for SOM ligands, which inevitably restricts the related applications such as protein-ligand complex structure prediction and virtual screening. 
As these two fundamental tasks have heavily relied on docking tools in the past four decades\cite{kuntz1982geometric,lengauer1996computational,luo2019challenges,pagadala2017software}, developing a deep learning model that provides refined atomic complex structure and screening for the protein-ligand pairs is highly demanded.

To this end, we introduce \ourmodelnoblank, a novel docking-free geometric deep learning method to accurately predict the native-like conformation of ligands with their corresponding protein targets and the respective binding strengths in one step (Fig. \ref{fig:framework}\textbf{c}). Specifically, for a given protein and ligand pair, the ligand and its binding target are jointly represented as a complete undirected graph, with each node denoting an atom, and all nodes are mutually connected. Then, their 3-D structures are directly optimized by the atom coordinates in the Euclidean space, with the binding strength and the correlation-enhanced graph learning jointly learned as auxiliary tasks (Fig. \ref{fig:framework}\textbf{d}). 

Based on the ability to represent the inter-atom relationships, \ourmodel achieved state-of-the-art performance on major tasks in drug development, \ie, structure prediction, virtual screening, and affinity estimation. Notably, \ourmodel outperformed $12$ existing docking tools with a significant improvement of $14\%$ in terms of success rate on conformation prediction, with an up to $1851$x faster inference speed. Moreover, our method performs even better on the more challenging and practical tasks, \ie, cross-docking and virtual screening, with greatly higher improvements of $20.1\%$ and $19.3\%$ in terms of success rate, respectively, compared with the existing best-performing method.
Evaluations on SARS-CoV-2 \mpro complex structure prediction show \ourmodel obtained $18.2\%$ improvements compared with docking tools, revealing its robustness in real environments. 
Despite the high accuracy and speed, the qualitative analysis showed that \ourmodel has the interpretability to learn non-covalent interactions even without explicitly incorporating the physical or chemical priors, \eg, the energies, which are usually regarded as the core idea of building structure prediction methods. 
The superior performance of \ourmodel suggests a promising paradigm for structure-based drug design using deep learning. We hope this work can facilitate the research of drug development.

\section{Results}\label{sec2}

\subsection{\ourmodel pipeline}\label{sec2.1}

\begin{figure}
\centering
\includegraphics[width=\linewidth]{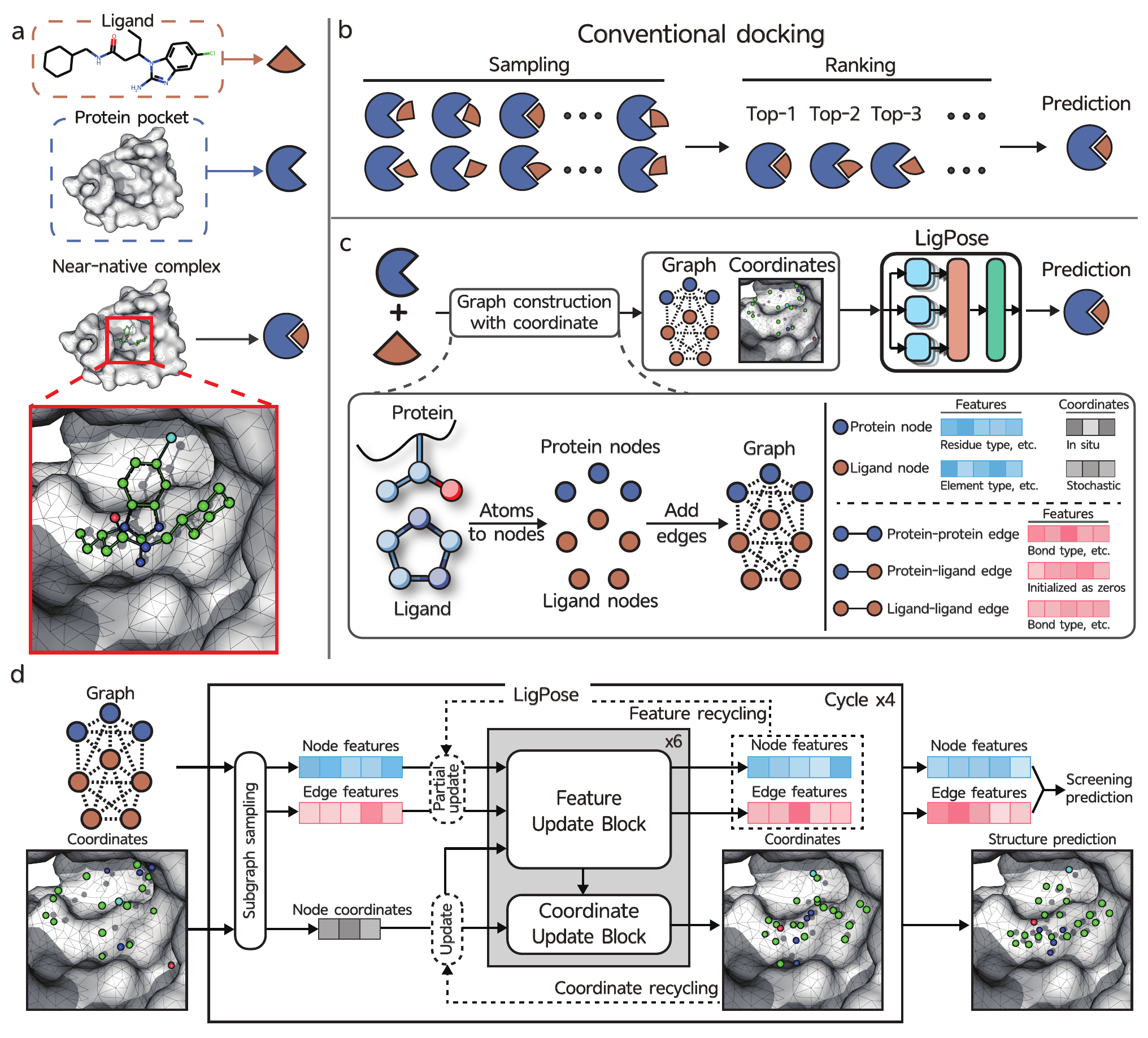}
\caption{\textbf{\ourmodel predicts ligand-protein complex structure using geometric deep learning in an end-to-end manner, compared with the conventional docking method.} (\textbf{a}) Notations of a ligand, a protein pocket, and their complex. (\textbf{b}) The pipeline of the conventional docking tool consists of two stages, \ie, sampling and ranking\cite{meng2011molecular}. (\textbf{c}) The pipeline of our method. (\textbf{d}) Architecture of \ourmodelnoblank.}
\label{fig:framework}
\end{figure}

\ourmodel predicts the 3-D structure of the ligand-binding conformation with its protein target in an end-to-end manner. As shown in Fig. \ref{fig:framework}\textbf{c}, a given protein and ligand are jointly represented by a complete undirected graph, where each atom is denoted as a node, representing its chemical feature and coordinate, and all nodes are mutually connected. Since the entire protein-ligand graph is large, \ourmodel (Fig. \ref{fig:framework}\textbf{d}) first adopts a sampling and recycling strategy, to predict with multiple cycles, where each cycle updates a randomly sampled sub-graph. 
The features and coordinates within the sub-graph are then processed by the proposed feature and coordinate update blocks. These two blocks are built on the graph neural network and stacked $6$ times with unshared weights. The key design of these two blocks is to leverage the inter-atom distances during network forwarding for maintaining the spatial information, making the predicted atom coordinates insensitive to their initial positions, but influenced by the inter-atom correlations. Finally, a symmetric-aware loss is proposed to optimize the network, combined with a stochastic coordinate initialization strategy for the ligands, to enable the method of distinguishing the ligand atoms with the same chemical features. The affinity and screening efficacy are learned by two additional prediction heads as auxiliary tasks. Moreover, we adopt self-supervised learning of the atomic correlations on large-scale unlabeled data including millions of random protein-ligand pairs without characteristics, to improve the generalizability of \ourmodel to unknown molecules. An ablation study of \ourmodel is illustrated in Suppl. Sec. \ref{Sec:ablation}. Please refer to Methods, Suppl. Methods and Fig. \ref{fig:coorupdate} for more details of the method. 

\subsection{Predicting accurate complex structures}\label{sec2.2}

\begin{figure}[htbp]
\centering
\includegraphics[width=0.97\linewidth]{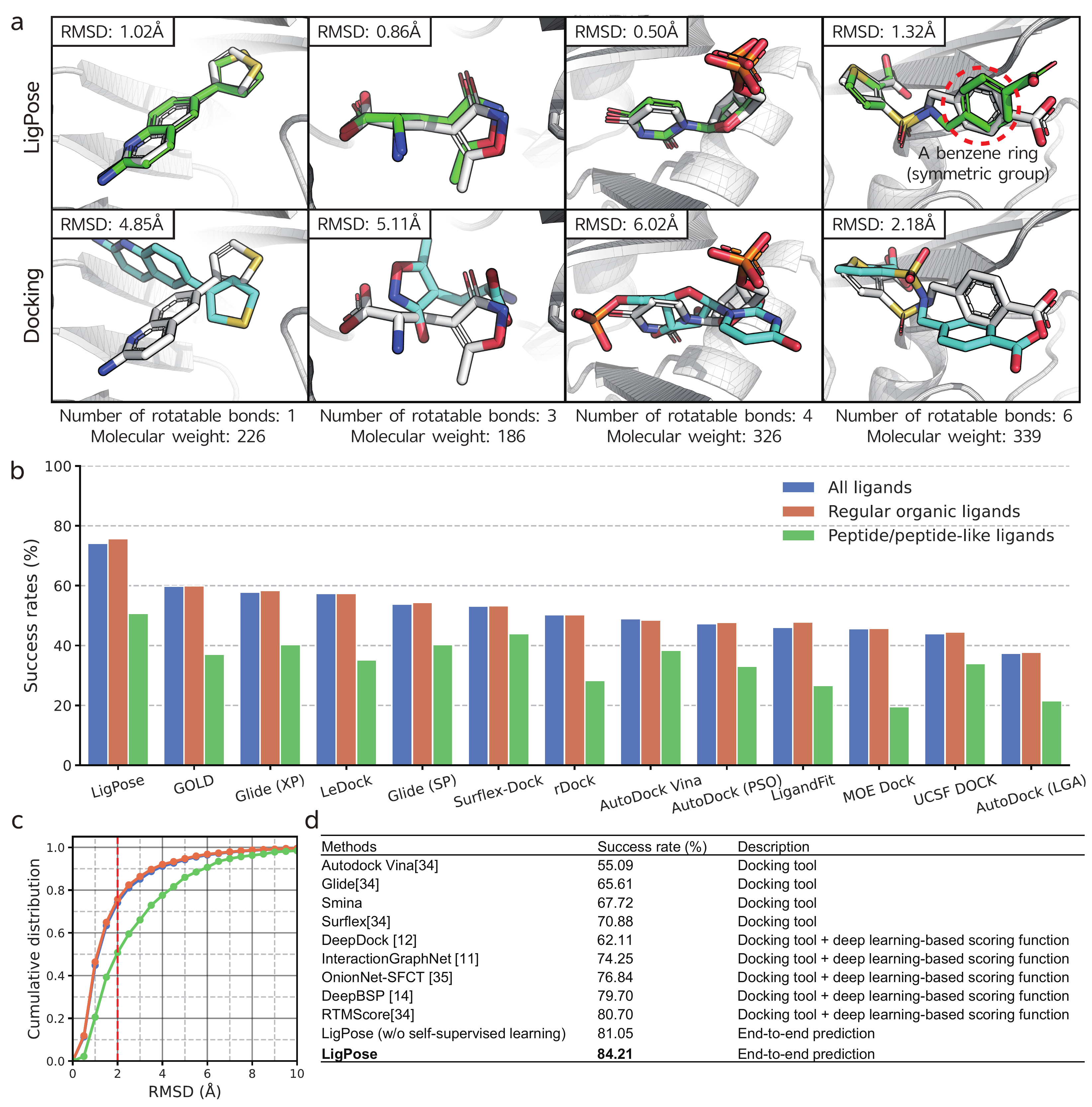}
\caption{\textbf{Performance of \ourmodel on ligand-binding conformation prediction compared with popular molecular docking tools and hybrid deep learning methods.} (\textbf{a}) Visualization of the generated poses of \ourmodel and a docking tool (Smina\cite{smina}) for four ligands with various weights and number of rotatable bonds (PDB codes: 3RSX, 1P1Q, 3DXG, 4JXS). The native poses, predictions of \ourmodelnoblank, and docking tools are denoted as white, green, and cyan backbones, respectively. Within the predictions, the oxygen and nitrogen atoms are denoted as red and blue colors, respectively. (\textbf{b}) Quantitative comparison of success rate between \ourmodel and the top-scored poses generated by $12$ docking tools on the refined set of PDBbind. (\textbf{c}) Cumulative distribution of RMSD of \ourmodelnoblank. The red dashed line indicates the RMSD threshold of 2\AA. Blue, orange, and green colors denote all ligands, the regular organic ligands, and the peptide/peptide-like ligands, respectively. (\textbf{d}) Quantitative comparison of success rate between \ourmodelnoblank, the docking tools, and the hybrid deep learning methods on the core set of PDBbind.}
\label{fig:betterthandocking}
\end{figure}

In experiments, we first focus on the flexible-ligand prediction with the determined protein pockets, a widely used setting in structure-based drug development\cite{huang2018comprehensive}, to demonstrate its effectiveness.
We compare \ourmodel with $12$ popular docking tools on the refined set of PDBbind database\cite{PDBbind,wang2016comprehensive}. The PDBbind database collects a large set ($N=19443$) of 3-D biomolecular complex structures from the PDB database. The refined set ($N=5316$) and the core set ($N=285$) are derived from PDBbind with high data quality regarding the crystal structures, binding data, and the nature of complexes\cite{li2014comparative}. 

The visualization of the predicted poses in Fig. \ref{fig:betterthandocking}\textbf{a} shows \ourmodel is able to predict accurate poses for ligands of various flexibilities on those that are failed by docking tools. It is worth noting that the symmetric structures (\ie, atoms in a benzene ring) are predicted in the right place. 
The quantitative comparison of the success rate between \ourmodel and $12$ popular docking tools with top-scored poses is shown in Fig. \ref{fig:betterthandocking}\textbf{b}. 
The predicted poses with Root Mean Square Deviation (RMSD) less than $2$\AA~to their native poses are regarded as successful predictions\cite{wang2016comprehensive}. As suggested by the figure, \ourmodel obtains a high success rate of $74.1\%$ on the entire refined set. Compared with a previous study that reported on a subset ($N=2002$) of the refined set\cite{wang2016comprehensive}, \ourmodel achieves $73.3\%$, which is $13.5\%$ higher than the best record docking tool (GOLD\cite{GOLD} with $59.8\%$). The other docking tools have unsatisfied success rates, ranging from $37.4\%$ (AutoDock (LGA)\cite{morris2009autodock4}) to $57.8\%$ (Glide (XP)\cite{friesner2004glide}). 
In addition, we observe the peptide/peptide-like molecules with more rotatable bonds and weights are more challenging to predict than the regular organic ligands, leading to performance degradation for both \ourmodel and docking tools. Considering the predicted poses with RMSD $<4$\AA~(Fig. \ref{fig:betterthandocking}\textbf{c}), \ourmodel achieves the success rate of above $90\%$ on regular organic molecules, shown effectiveness on near-native structure prediction for conformations.

We also report the performance of \ourmodel on regular organic molecules compared with the docking tools with the best poses in Suppl. Fig. \ref{fig:bestpose}. Herein, the best pose denotes the pose of the lowest RMSD among all sampled poses and reveals the performance upper bound of the docking tools. 
The figure shows the success rate of \ourmodel ($74.6\%$ and $58.4\%$) outperforms $8$ and $10$ docking tools for regular organic and peptide/peptide-like ligands, respectively. These results indicate that the majority of those docking tools could not have a chance to outperform \ourmodel by improving the scoring functions\cite{morrone2020combining}, since they cannot produce correct pose candidates in limited sampling steps for some ligands that are correctly predicted by \ourmodelnoblank. 

To conduct a fair comparison with the deep learning-based methods, we performed \ourmodel on the core set, a well-designed but smaller subset of PDBbind. Here we summarized the success rates of \ourmodel with diversified methods, including docking tools and state-of-the-art deep learning-based methods\cite{DeepBSP,IGN,mendez2021geometric,RTMScore,zheng2022improving} in Fig. \ref{fig:betterthandocking}\textbf{d}, where \ourmodel obtained the success rate of $84.2\%$, which is higher than the best-performing method RTMScore ($80.7\%$), indicating the high performance of the proposed end-to-end paradigm. Besides, we extend \ourmodel to work as a scoring function (see Suppl. Results Sec. \ref{result_score} and Fig. \ref{fig_score}) to make it compatible with the existing docking tools, which also achieves comparable results to the state-of-the-art scoring functions.


We further test \ourmodel on cross-docking, a more challenging and practical task in drug screening\cite{tuccinardi2014extensive,shen2021impact}. Cross-docking aims to produce the binding conformation for a ligand, from which the target protein is experimentally determined by a complex with another ligand\cite{tuccinardi2014extensive}. The experiments are implemented on the PDBbind-CrossDocked-Core set ($N=1343$ pairs, derived from the core set\cite{shen2021impact}), where \ourmodel achieves the highest success rate of $72.0\%$, with a greatly large improvement ($20.1\%$) compared to the second-best method (\ie, RTMScore, $51.9\%$), as shown in Suppl. Table \ref{tablecross}, revealing a more generalizable method of \ourmodel compared with existing competitors. Moreover, the success rate of \ourmodel shows improvements of $5.3\%$ to $20.1\%$, compared with three docking tools using the best-sampled poses ($51.9\%$ to $66.7\%$), suggesting the advances of the data-driven approach. 

\subsection{Ligands with various flexibilities}\label{sec2.5}

\begin{figure}[htbp]
\centering
\includegraphics[width=\textwidth]{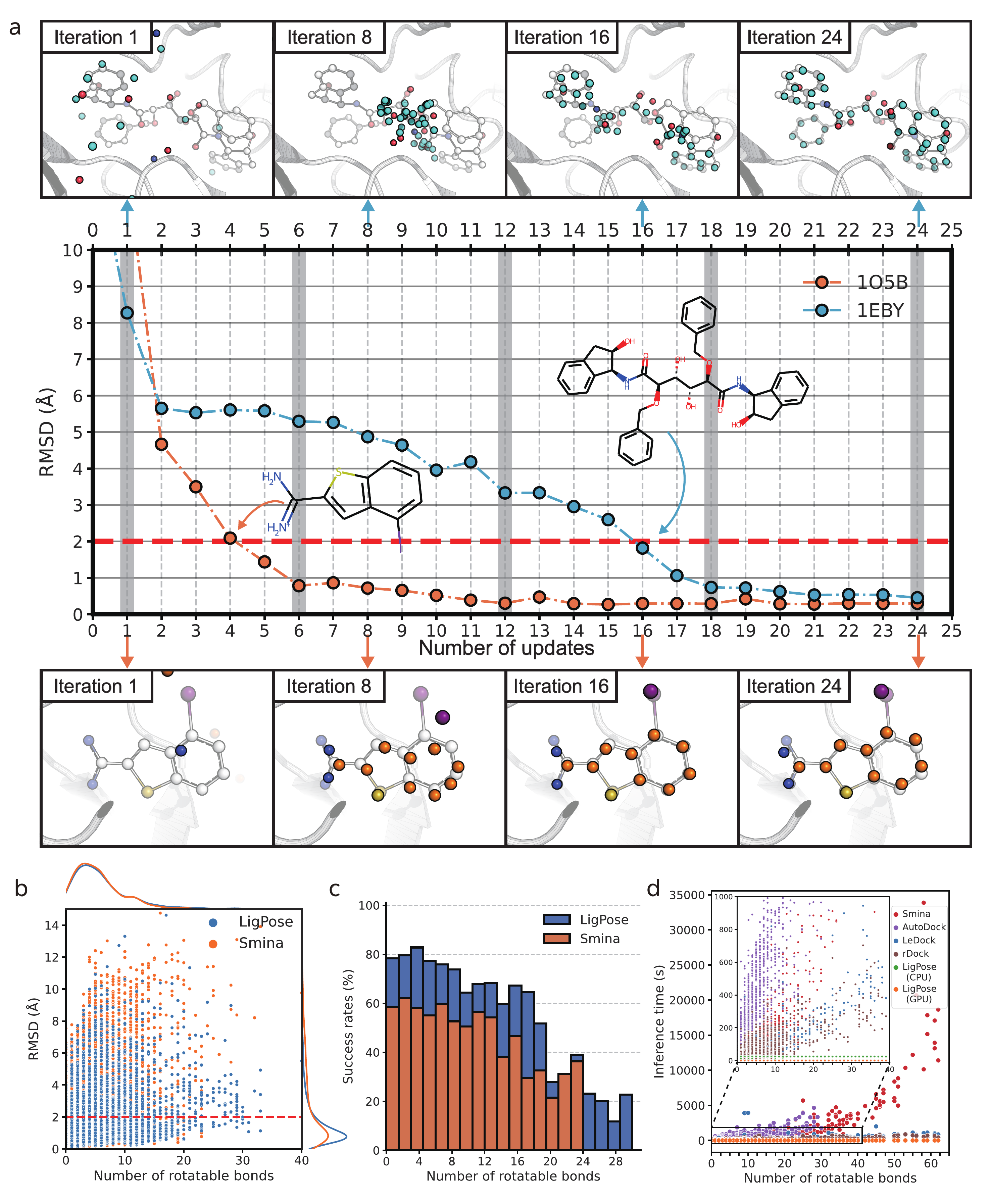} 
\caption{\textbf{Performance of \ourmodel with respect to the ligand flexibility.} (\textbf{a}) RMSD trajectory of two representative samples (PDB codes: 1O5B and 1EBY). \ourmodel updates a given ligand $24$ times using $4$ cycles, with each cycle updating the ligand $6$ times. The ligand atoms in the $\{1,8,16,24\}$th updates are visualized at the panel's top and bottom. The predicted atoms are denoted with orange and cyan colors for 1O5B and 1EBY, respectively. The oxygen, nitrogen, and iodine atoms are denoted with red, blue, and purple colors, respectively. The native poses are placed in the background with a grey color. (\textbf{b-c}) RMSD (\textbf{b}) and success rate (\textbf{c}) for ligands with respect to the number of rotatable bonds on the core set of PDBbind. The red dashed line indicates the RMSD threshold of 2\AA. (\textbf{d}) Inference time of \ourmodel and four popular docking tools on $1000$ randomly selected samples in the PDBbind dataset.}
\label{fig:hidden}
\end{figure}

To solve the ligands with high flexibilities (\ie, with many rotatable bonds), the strategies of sub-graph sampling and recycling are combined to progressively refine the predictions. We visualize the ligand atoms and plot the RMSD trajectories for two representative samples in Fig. \ref{fig:hidden}\textbf{a}, where one is heavier with more rotatable bonds (PDB code: 1EBY) than the other (PDB code: 1O5B). As a hard example, the prediction of 1EBY reaches the threshold of 2\AA~after $16$ updates, which is much slower than that of 5O87 (with $5$ updates), indicating the effectiveness of the proposed recycling strategy.

To quantify the performance of \ourmodel concerning the number of rotatable bonds, we plot the results of \ourmodel and Smina on the refined set in terms of RMSD (\textbf{b}) and success rate (\textbf{c}) in Fig. \ref{fig:hidden}. One can observe is the difficulty of prediction is positively correlated with the number of rotatable bonds. We suppose the reason is that more rotatable bonds indicate higher flexibility of the ligand. Although the success rates of \ourmodel and Smina are both decreased when predicting ligands with many rotatable bonds, \ourmodel performed consistently better than Smina through the dataset. 
Moreover, Smina failed to predict with more than $24$ rotatable bonds, indicating the limited processing power of the docking methods. By contrast, \ourmodel correctly predicted $20\%$ of them, showing the merit of our learning-based methodology. Similar findings were discovered from the core set, as shown in Suppl. Fig. \ref{fig:corerotbond}. 

The efficiency is also a core factor of drug development\cite{mayr2009novel}, caused by a large number of drug-like molecules for screening and measurement. We then plot the inference time per ligand with respect to the number of ligand rotatable bonds for \ourmodel and Smina in Fig. \ref{fig:hidden}\textbf{d}. 
For a fair comparison, we use a single CPU core to implement docking tools and \ourmodel (CPU), with only an additional common GPU device (Nvidia GTX 2080Ti) for the GPU version (\ourmodel (GPU)). 
From the figure, we observed the inference time of \ourmodel is constant, in contrast to that of docking tools, which is positively correlated with the number of rotatable bonds. On average, \ourmodel inferences $4-26$ (CPU) / $303-1851$ (GPU) times faster than docking tools. For ligands with more than $10$ rotatable bonds, \ourmodel can infer $7299$x times faster than docking tools, and up to $14256$x faster for ligands with more than $20$ rotatable bonds.



\subsection{Accurate and fast screening}


\begin{figure}[H]
\centering
\includegraphics[width=\textwidth]{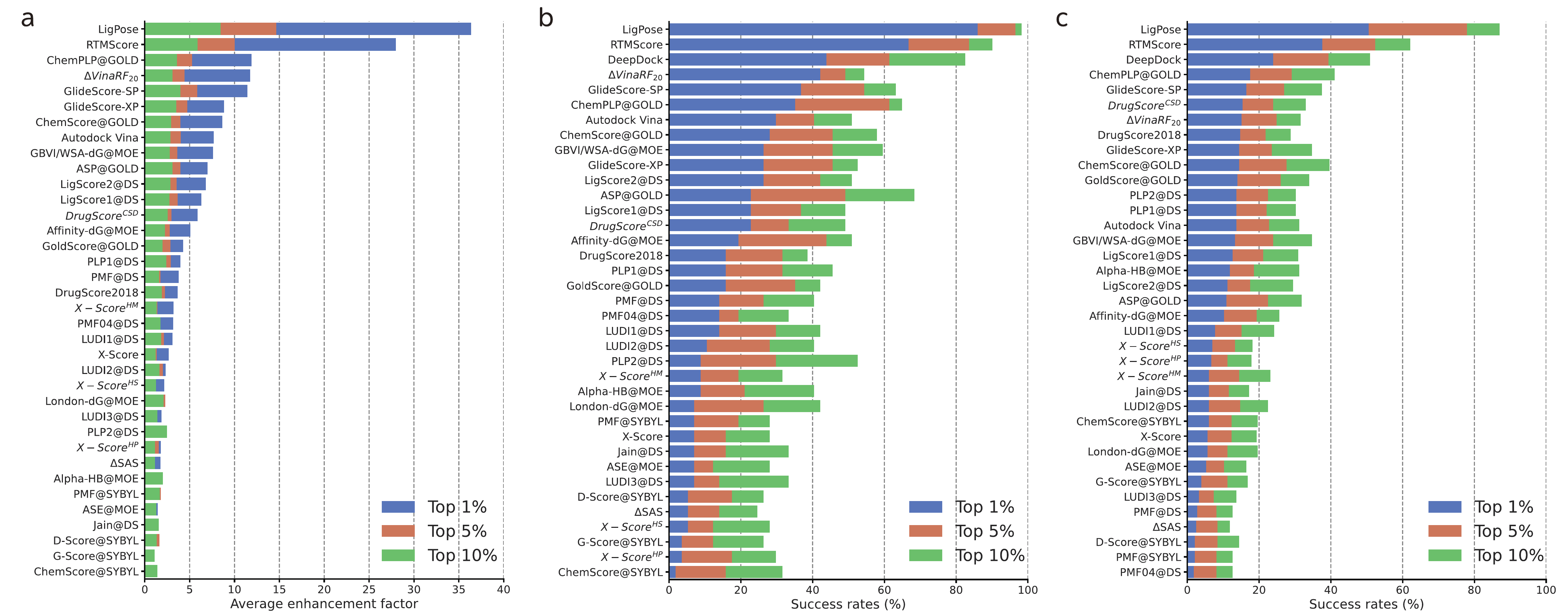} 
\caption{\textbf{Screening power of \ourmodel on the CASF-2016 benchmark.} 
(\textbf{a}) Average enhancement factors of forward screening.
(\textbf{b}) Success rates of forward screening. 
(\textbf{c}) Success rates of reverse screening.}
\label{fig:vs}
\end{figure}

Accurate prediction of the complex structure is an essential procedure, and ideally beneficial for the virtual screening task. 
Therefore, we validate the screening power of \ourmodel using the CASF-2016 benchmark, which contains $57$ proteins, each having at least $5$ true binders with a wide range of affinities. 
Therefore, for each SOM-protein pocket pair, \ourmodel predicts the potential binding strength of SOMs to the protein pockets as the screening score, through the multiplication of the predicted binding probability and affinity. 

As suggested by Fig. \ref{fig:vs}, \ourmodel outperformed the state-of-the-art methods in all three metrics with a very large performance gap, compared with recent deep learning-based methods and popular conventional methods. 
Concretely, for the forward screening, \ie, the task of identifying true binding SOMs for a certain protein, \ourmodel gets an average enhancement factor (EF) of $36.4\%$, and a success rate of $86.0\%$ on top $1\%$-ranked SOMs. These results are greatly higher than that of the second best performing method RTMScore\cite{RTMScore} (with EF of $28.0\%$ and success rate of $66.7\%$), with improvements of $8.4\%$ and $19.3\%$ on EF and success rate, respectively. 
It also largely outperforms recent deep learning-based methods, \ie, DeepDock\cite{mendez2021geometric} (with EF of $16.4\%$ and success rate of $43.9\%$) and PIGNet\cite{moon2022pignet} (with EF of $19.36\%$ and success rate of $55.4\%$). Besides, the results show huge gaps to conventional methods (with EFs ranging from $0.8\%$ to $11.9\%$ and success rates ranging from $1.8\%$ to $42.1\%$). For top-$5\%$ and top-$10\%$ ranked SOMs, similar conclusions can be made. Notably, \ourmodel reaches very high success rates of $96.5\%$ and $98.2\%$ on top-$5\%$ and top-$10\%$ ranked SOMs, suggesting that nearly all true SOM binders are settled at the top of the predictions. 
Similarly, for the reverse screening, \ie, the task of identifying true binding proteins for a certain SOM, \ourmodel obtains success rates of $50.5\%$, $77.9\%$, and $87.0\%$ in top-$1\%$, top-$5\%$, and top-$10\%$ ranked proteins, respectively. It also outperforms RTMScore\cite{RTMScore} (with $37.6\%$, $52.4\%$, $62.1\%$) and DeepDock\cite{mendez2021geometric} (with $23.9\%$, $39.3\%$, $50.9\%$) by a large margin.

The superior screening ability of \ourmodel benefits from its state-of-the-art performance on drug-target affinity estimation (see details described in Suppl. Results Sec. \ref{result_aff} and Table \ref{table_aff}). Notably, \ourmodel provides accurate affinity estimation without requiring native complex structures. 

\subsection{Validating on the SARS-CoV-2 \mpro} 

\begin{figure}[htbp]
\centering
\includegraphics[width=0.97\textwidth]{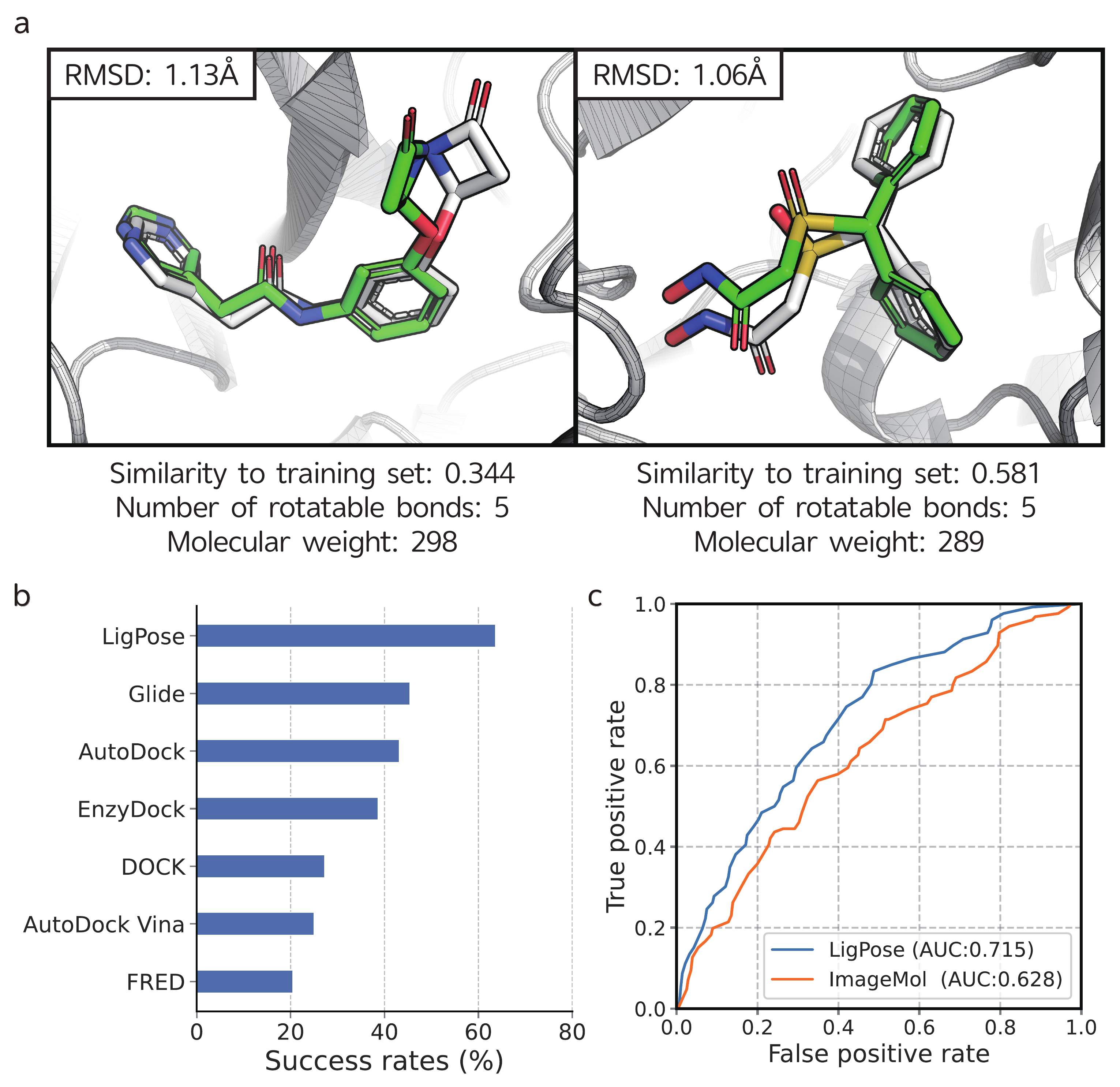}
\caption{\textbf{Applications of \ourmodel on drug research for SARS-CoV-2 \mpronoblank.}
(\textbf{a}) Two visualized samples of \mpro complexes (PDB codes: 5RGU and 7ANS). Predictions of \ourmodel are denoted as green and cyan backbones. Within the predictions, the oxygen, nitrogen, and sulfur atoms are denoted as red, blue, and yellow colors, respectively.
(\textbf{b}) Success rates of structure prediction for \mpro complexes. 
(\textbf{c}) Success rates of virtual screening for \mpro inhibitors.}
\label{fig:Mpro}
\end{figure}

We further validate \ourmodel on the main protease (\mpronoblank) of severe acute respiratory syndrome coronavirus 2 (SARS-CoV-2) to demonstrate its efficacy in real-world applications. \mpro plays a critical role in virus replication as a hot-spot target for anti-SARS-CoV-2 therapy\cite{gil2020covid}. In this work, two tasks are considered, \ie, structure prediction and inhibitor discovery. 

The structure prediction ability is assessed on a recently developed benchmark, which provides a baseline of $6$ commonly-used docking tools on $44$ \mpro complex structures\cite{zev2021benchmarking}. We visualize two typical samples in Fig. \ref{fige:Mpro}\textbf{a}, where \ourmodel successfully predicts binding conformations of these novel inhibitors (Tanimoto similarities\cite{bajusz2015tanimoto} of $0.344$ and $0.581$ to their most similar molecules in the training set using Morgan fingerprint\cite{rogers2010extended}). 
For the quantitative comparison of all $44$ complexes as shown in Fig. \ref{fig:Mpro}\textbf{b}, \ourmodel achieves a success rate of $63.64\%$ with a significant improvement of $18.19\%$, compared with the best-performing docking tool (\ie, Glide, $45.45\%$). The success rates of other docking tools are much lower in the range of $20.45\%$ to $43.18\%$. 

We validate \ourmodel on screening \mpro inhibitors on a dataset ($N=344$) derived from DrugCentral\cite{kc2021machine}. The results in Fig. \ref{fig:Mpro}\textbf{c} show \ourmodel obtains an Area Under Curve (AUC) of $71.5\%$, with an improvement of $8.7\%$ compared with the recent deep learning-based methods ImageMol ($62.8\%$)\cite{zeng2022accurate}.

\subsubsection{Learning non-covalent interactions}\label{result_noncov}
\begin{figure}
\centering
\includegraphics[width=\linewidth]{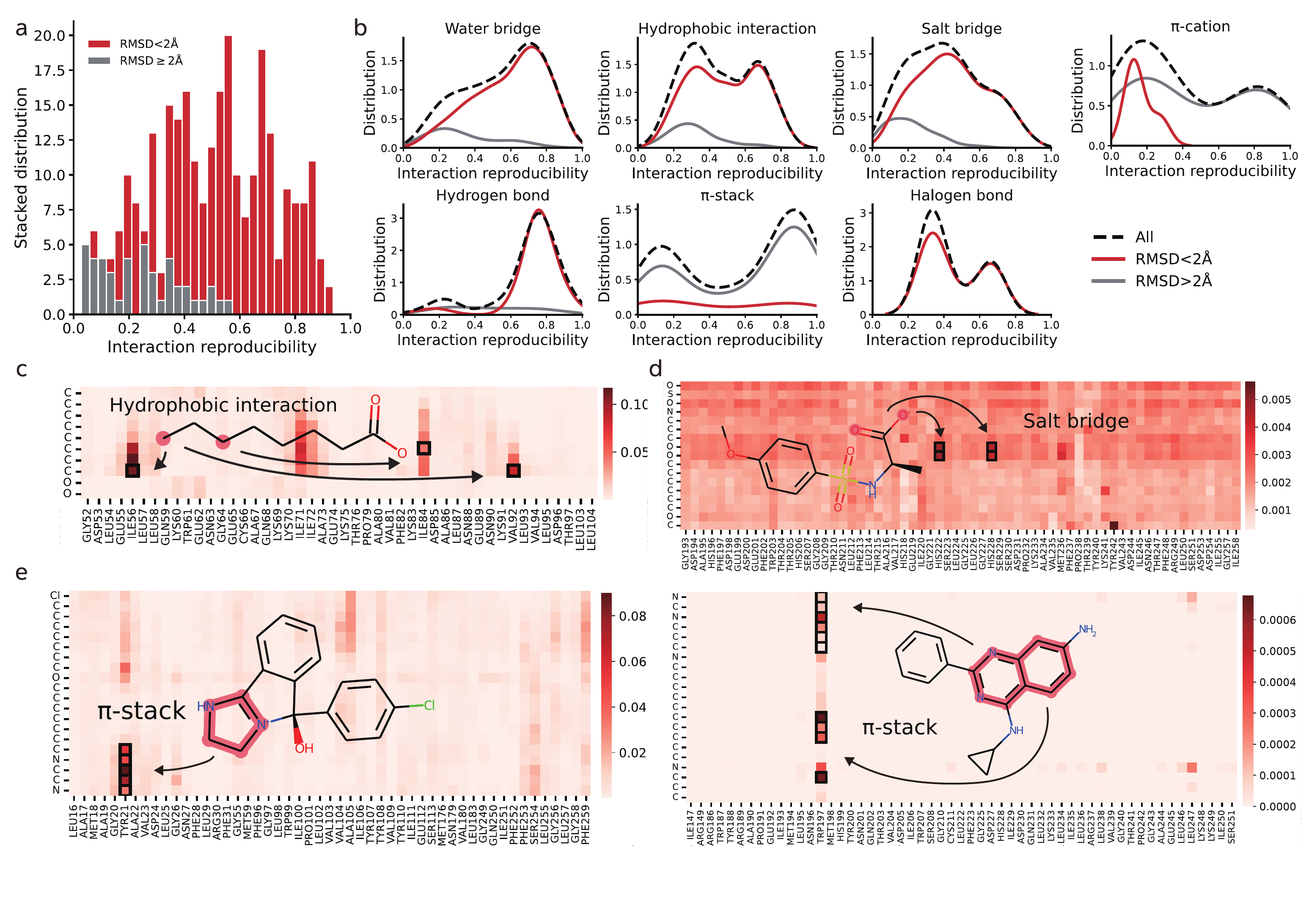}
\caption{\ourmodel implicitly learned non-covalent interaction patterns. (\textbf{a}) Stacked distribution of interaction reproducibility for all non-covalent interactions on the refined set. The red and grey colors denote RMSD\textless2\AA~and RMSD$\ge$2\AA, respectively. (\textbf{b}) Distribution of the interaction reproducibility for seven specific non-covalent interactions. Distribution of all samples, the samples of RMSD\textless2\AA~ and RMSD$\ge$2\AA~ are denoted in black, red, and grey dashed lines, respectively. (\textbf{c-f}) Visualization of attention weights in \ourmodel for three representative protein-ligand complexes. The attention weights with the red color from light to deep indicate the probability of non-covalent interaction between a ligand atom and a protein residue from low to high. Black boxes denote non-covalent interactions of native structures in the attention maps for hydrophobic interactions (PDB code: 3NQ9) (\textbf{c}), Salt bridge (PDB code: 3EHY) (\textbf{d}), and $\pi$-stack (PDB code: 4MME/4DLI) (\textbf{e,f}). Attention weights are visualized at the atom level for ligands and residue level for proteins.}
\label{fig_noncov}
\end{figure}

In addition to the effectiveness and efficiency, \ourmodel also showed good interpretability by investigating its ability to reconstruct the non-covalent interactions.
We used a metric, namely interaction reproducibility, to quantify this ability, following the workin \cite{paggi2021leveraging} (see Methods for details). 

From Fig. \ref{fig_noncov}\textbf{a}, we observed the interactions are well reconstructed by \ourmodel on predictions with RMSD $\textless 2$\AA, and have lower scores on predictions with RMSD$\ge 2$\AA. 
For each type of interaction (Fig. \ref{fig_noncov}\textbf{b}), \ourmodel performed best for hydrogen bond and water bridge and showed insensitive to $\pi$-cation and halogen bond. We suppose the reason is the lack of respective ligand samples. We further compared the performance of \ourmodel with Smina (Suppl. Figs. \ref{fig:dockingnoncov1} and \ref{fig:dockingnoncov2}), and found that \ourmodel showed consistently better performance for all the seven types of interactions. 

In addition to the high quantitative scores, we further inspected the attention weights, by visualizing three representative complexes with respect to three non-covalent interactions (\ie, hydrophobic interactions, salt bridge, and $\pi$-stack) in Fig. \ref{fig_noncov}(\textbf{c-f}). We observed from the figure that \ourmodel can capture some potential interactions between ligands and proteins, without explicitly using related physical or chemical prior knowledge.

\section{Discussion}\label{sec3}

In this work, we propose \ourmodelnoblank, which can predict protein-ligand complex structure, affinity, and screening probability simultaneously using multi-task geometric deep learning. The protein-ligand pair is first presented by one complete undirected graph, then, \ourmodel with a graph transformer network structure, directly predicting the complex structure in three-dimensional space. To precisely predict the complex structure with atomic accuracy, \ourmodel is equipped with three innovations. To be specific, 1) we use a sampling and recycling strategy to optimally learn the protein information by multiple cycles, to handle a large number of protein nodes while reducing the computation burden. In each cycle, the feature update block is used to extract latent features for graph nodes and edges, and the coordinate update block updates the coordinates of all atoms (nodes) to their near-native 3-D position. Besides, 2) we design a specific atom initialization strategy and a symmetric-aware loss function for symmetric structures of the ligands. 3) We adopt a multi-task learning framework to train \ourmodel on all tasks, and leverage the power of self-supervised learning on large-scale unlabeled protein-ligand pairs, making the model more generalized to novel molecules. These advances obtained a notable improvement in the tasks of drug development.  


\ourmodel correctly predicted $3914$ ($74\%$) ligand structures (with RMSD $<2$\AA) on the refined set of PDBbind, which significantly outperformed $12$ popular docking tools by $14\%$. \ourmodel also performed better than recent hybrid deep learning methods on the core set. Many pieces of research in this field focused on building hybrid methods\cite{lim2019predicting,morrone2020combining,DeepBSP,IGN,yang2021deep,ragoza2017protein}. However, as we demonstrated in experiments (Fig. \ref{fig:betterthandocking}\textbf{e}), these methods are limited by the sampling power of the docking tools\cite{morrone2020combining}. Indeed, exhaustively exploring the sampling space dramatically increases the runtime, making it impractical in real-world drug screening. 
Recently, Oscar \etal\cite{mendez2021geometric} proposed DeepDock, which used geometric deep learning to build a docking-like method. Interestingly, this method could be regarded as a deep learning-based scoring function from a global perspective. However, this method in a docking-like fashion will also face the same challenges as docking tools in both effectiveness and efficiency. For instance, they reported that $60$ out of $285$ samples failed to find a minimum, and only $62\%$ ($177$) out of $285$ samples were successfully predicted on the core set, leading to unsatisfied performance even worse than conventional docking methods (Fig. \ref{fig:betterthandocking}\textbf{e}). 
By contrast, \ourmodel overcomes these limitations by directly optimizing the atom coordinates without sampling and energy-based minimizing, suggesting a novel computational paradigm for the prediction of the protein-ligand complex structures, coincident with AlphaFold in the task of protein structure prediction\cite{AF2}. 
In a more challenging task, \ie, cross-docking, \ourmodel shows tremendous success with an improvement of $20.1\%$ in terms of success rate compared with the best-recording deep learning method RTMScore. Unlike the commonly used re-docking task 
adopted by previous methods (\eg, DeepDock, DeepBSP), cross-docking is more practical in real-world situations.


In addition to the high performance, \ourmodel also inferences considerably faster than docking.
In real-world drug development, methods are expected to accurately screen a tremendous number (\ie, up to $10^{60}$\cite{bohacek1996art}) of drug-like molecule candidates\cite{lyu2019ultra,sadybekov2021synthon}.
Unfortunately, the docking-based methods with an iterative sampling strategy have suffered from the efficiency problem for a long time. 
Many studies tried to solve the problem by implementing CPU/GPU parallel acceleration algorithms in favor of engineering\cite{fan2021gpu,ding2020accelerated,solis2022benchmarking}.
Besides, some works combined deep learning with conventional docking tools to boost the inference speed\cite{deepdocking,Emulating}. 
We suggest that \ourmodel with an efficient end-to-end learning framework could be an alternative choice for docking to accelerate drug development. 

Precise structure prediction naturally promotes drug screening, which is dominated by docking in large-scale structure-based drug development. Thus, structure-based deep learning methods\cite{deepdocking,ma2021structure,RTMScore} also prefer incorporating the compute-intensive docking tools, which limit their accuracy and efficiency. Other than the structure-based methods, several recently developed deep learning methods, including image-based (\ie, using molecule image)\cite{zeng2022accurate}, graph-based (\ie, using molecular graph)\cite{wang2022pre,pham2021deep,fang2022geometry,li2021effective}, and sequence-based (\ie, using Simplified molecular input line entry system (SMILES))\cite{guo2022improving,skinnider2021chemical,wang2021multi}, learn informative representations of SOMs, showing promising performance in the binary classification of binding candidates. However, these ligand-based methods lack consideration for protein structure (or sequence), resulting in providing a black box method without actual atomic interactions between proteins and ligands. \ourmodel provides a more reasonable solution to combine the structure basis and protein-ligand interactions for drug screening. Therefore, \ourmodel achieves the state-of-the-art affinity estimation and virtual screening performance without leveraging the native complex structures, showing its feasibility to real applications. 
Our further validations on SARS-CoV-2 \mpro structure prediction and inhibitor screening indicate that \ourmodel meets the requirements for real drug development.



Building an accurate automatic drug development pipeline has encouraged decades of exploration\cite{ekins2019exploiting,baek2022deep}. Recent advances demonstrated that protein structures can be accurately predicted by AlphaFold\cite{AF2} and RoseTTAFold\cite{RoseTTAFold}. Therefore, predicting the structure of biomolecular complexes could be naturally considered as a subsequent application, as also suggested by Baek \etal\cite{baek2022deep}. 
In addition, the task of identifying the protein pocket has been well-addressed by existing works\cite{abdollahiresidue,gainza2020deciphering,pu2019deepdrug3d,aggarwal2021deeppocket}. 
Inspired by these successes, integrating \ourmodel with the aforementioned methods is appropriate, carrying out a promising end-to-end pipeline of drug screening directly from the protein sequence and the ligand chemical formula.

Moreover, in the field of structural biology, recent attempts have extended the application scenario of artificial intelligence to design the protein structures\cite{huang2022backbone,anishchenko2021novo}. A similar idea could also be applied to drug development. However, existing methods proposed to design (or generate) novel drug candidates neglected to consider either the 3-D structures or the protein-drug interactions\cite{wang2022molecular,popova2018deep,wang2022comprehensive,li2021effective,walters2020applications,bai2021molaical,wang2021deepr2cov,krishnan2021accelerating,moret2021beam}. 
Here, we suggest that \ourmodel has the potential to build novel end-to-end drug-generation methods with both refined structural details and confident bioactivity prediction, to boost de novo drug design, which can be regarded as future work. 


\clearpage

\section{Methods}\label{sec4}

\subsection{Data collection}

\subsubsection{Benchmark Dataset}

\noindent\textbf{Training and validation dataset.} We use the \emph{general set} of PDBbind database (version 2020)\cite{PDBbind} to develop and validate \ourmodelnoblank, consisting native structures of $19443$ protein-ligand complexes. Within the general set, two subsets are used for evaluation, \ie, the \emph{refined set} and the \emph{core set}, which are of better data quality for evaluating docking tools and scoring functions in existing works. We used the refined set to validate the performance of \ourmodel with five-fold cross-validation. In each fold, the rest of the samples in the general set with non-overlapping to the validation set are used for training. Performance for $2002$ selected complexes in the refined set as adopted in \cite{wang2016comprehensive} under a similar setting is also reported. For the core set we obtain the training data following the previous deep learning-based methods, \eg, DeepDock\cite{mendez2021geometric}, RTMScore\cite{RTMScore}.

The peptide-like SOMs are identified by the Biologically Interesting Molecule Reference Dictionary (BIRD)\cite{BIRD}. 
$2200$ random selected samples in the training set are used for hyper-parameter search, and the selected architecture is used for all tasks. 
The unsuccessfully processed data by RDKit\cite{rdkit} is removed. None of the samples in the core set but $14$ samples in the refined set failed to be processed. Finally, only a minor proportion of the entire PDBbind database ($153, <1\%$) is dropped. 

\noindent\textbf{Auxiliary unlabeled dataset.} 
The structure-determined complexes are only a tiny portion of the complex family. To enhance the performance and generalizability of \ourmodelnoblank, we add a task to \ourmodel to train it under a self-supervised learning scheme, which models the data structures from large-scale unlabeled data, \ie, with unknown complex structures and protein/ligand properties. To this end, we collected a large unlabelled dataset, including a total number of $2147477$ SOMs and $171789$ proteins.
The SOMs are derived from two databases, \ie, DrugBank\cite{wishart2018drugbank} ($11290$) and ChEMBL\cite{gaulton2012chembl} ($2136187$), with diversified drug/drug-like compounds that are widely used for the pre-training of SOM features\cite{xia2022pre}. The proteins are derived from the PDB database\cite{berman2000protein}. 

\noindent\textbf{Test dataset.} The whole core set and the benchmark dataset CASF-2016\cite{su2018comparative} derived from the core set with carefully designed poses are used for testing. We adopt four main tasks, \ie, scoring power (the ability to estimate binding affinity), ranking power (the ability to identify near-native pose), docking power (the ability to identify near-native pose), forward screening power (the ability to identify potential ligands for a target protein) and reverse screening power (the ability to identify potential target proteins for a SOM), to comprehensively assess the performance of \ourmodelnoblank. As abovementioned, all complexes in the core set are removed from the training set to make them non-overlapped.

\subsubsection{Real-world dataset}

To demonstrate the ability of \ourmodel in real applications, we perform it on two major tasks (\ie, complex structure prediction and screening) for the drug development of SARS-CoV-2 \mpronoblank.

\noindent\textbf{\mpro structure dataset.}
We adopt the dataset as reported in \cite{zev2021benchmarking}, which contains $44$ non-covalent and non-surface-bound ligands with \mpro from SARS-CoV-2.

\noindent\textbf{\mpro screening dataset.}
The screening set ($N=344$) contains SOM binding data by experimentally measuring the enzymatic activity of \mpronoblank, which is collected from DrugCentral\cite{kc2021machine}, followed with a processing protocol introduced in ImageMol\cite{zeng2022accurate}.

\subsection{Pre-processing}\label{preprocess}

In this work, the protein pocket was adopted as the binding target, since obtaining the atomically refined structure is the major concern in drug development\cite{bender2021practical,kitchen2004docking}. Besides, many methods have been successfully developed to predict the ligand-binding pockets with high accuracies\cite{abdollahiresidue,gainza2020deciphering,pu2019deepdrug3d,aggarwal2021deeppocket,halgren2009identifying,laurie2005q}.

The protein pockets are the surrounding amino acids of the ligands in 3-D space. For the unlabeled data, we complete the missing atoms with modeler\cite{eswar2008protein} and search the pockets using the Fpocket\cite{le2009fpocket} method with the default setting. All pockets are filtered with a Druggability Score of $>0.5$, therefore, $631687$ pockets are finally used. Fpocket recognizes the pocket and represents the pocket center as virtual atoms. A residue is selected as the part of a pocket when its C$\alpha$ atom is within $13$\AA~to the nearest virtual atom. For the labeled data, a residue is selected as the part of a pocket when its C$\alpha$ atom is within $15$\AA~to the nearest ligand atom. Water molecules are not considered in this study.

For a given protein-ligand pair, \ie, a random pair in the unlabeled dataset or a native complex in the PDBbind dataset, a complete undirected graph is firstly constructed using RDKit\cite{rdkit}, with each node representing an atom, and all nodes (including themselves) are mutually connected. We initialize the graph by features listed in Suppl. Table \ref{table_feature}. In particular, the nodes are initialized as vectors with a length of $79$ for a protein and $45$ for a ligand, containing the chemical features. Edges are also initialized as vectors with a length of $7$ including covalent bond features and distance. Distances connecting different rigid parts are masked with $-1$. To enable the ability of SE(3)-equivalence for the method, and also augment the data, we implement a stochastic strategy to initialize the coordinates of the ligands. In this case, the protein nodes are kept in their original positions, while the positions of the ligand nodes are randomly initialized inside the pocket with an empirical distance. In this work, we initialize them in a Gaussian distribution with a standard deviation of $10$\AA.

\subsection{\ourmodel architecture}\label{sec4.3}

\ourmodel directly predicts the structure of the ligand-binding conformations with their target protein pockets in the 3-D space in an end-to-end manner. As shown in Fig. \ref{fig:framework}\textbf{c}, \ourmodel has a graph transformer architecture, with three major components, (1) a sampling and recycling strategy, (2) a feature update block followed by a coordinate update block to forward the features and coordinates through the graph, and the two blocks are stacked $6$ times with unshared weights, and (3) a stochastic initialization method for ligand nodes and a novel symmetric-aware loss. We will then illustrate them below.

\subsubsection{Sampling and recycling}\label{sec4.3.1}

Typically, a protein pocket contains hundreds of atoms. Using all of them as the input is inefficient and memory unaffordable for common GPU devices. Therefore, we adopt a sampling strategy to generate a sub-graph consisting of all \emph{core atoms} and some randomly selected \emph{context atoms}, with their respective edges, to feed to the network. Specifically, the core atoms involve all ligand atoms, and the C$\alpha$, C$\beta$ atoms of the protein, as they are enough to determine the position and orientation of the amino acids and protein backbone\cite{gront2007backbone,kmiecik2016coarse}. The other protein atoms are regarded as the context atoms to describe the structural details of the amino acids. Since the nodes are not fully used, we further introduce a recycling strategy, to enhance the representability of the method, as inspired by \cite{AF2}. In each cycle, a new sub-graph is sampled and forwarded, then, the updated graph is reused in the next cycle. In the next cycle, the newly sampled graph directly inherits the coordinates of corresponding nodes from the last cycle, and its features of the core atoms are combined with the last updated features by using an element-wise gate on the new features, implementing a partial update. (See details in Suppl. Sec. \ref{sec5.1.2})

\subsubsection{Feature Update Block}\label{sec4.3.2}

The feature update block maintains a graph Transformer-based architecture to forward the node and edge features. 
It iteratively performs two operations, \ie, message aggregation and feedforward. 
In message aggregation, information is aggregated from neighboring nodes to a central node using the multi-head attention (MHA) mechanism (Fig. \ref{fig:coorupdate})\cite{GAT}. In addition, the edge features are also incorporated to enhance the representation.

We use conventional cross-attention to calculate the MHA in the network, according to Query (denote as $\mathbf{q}$), Key (denote as $\mathbf{k}$), and Value (denote as $\mathbf{v}$)\cite{GAT} (See details in Suppl. Sec. \ref{sec5.1.2}).
Briefly, $\mathbf{q}$ and $\mathbf{v}$ are produced by the features of the central nodes and the neighboring nodes, respectively (see Fig. \ref{fig:coorupdate}\textbf{b}). $\mathbf{k}$ is extracted from the edge feature and a linearly transformed neighboring node feature (Fig. \ref{fig:coorupdate}\textbf{a-b}). The edge feature here is enhanced by the spatial information using the distance between the central node and its neighboring node (Fig. \ref{fig:coorupdate}\textbf{a}).
A softmax layer is then applied to the product of $\mathbf{q}$ and $\mathbf{k}$ to get attention masks for each $\mathbf{v}$. 
The central node aggregates the masked $\mathbf{v}$ to get an updated feature, and is followed by a linear layer to merge multi-head information (Fig. \ref{fig:coorupdate}\textbf{c}). An element-wise gate is implemented on these combined results.
Finally, a two-layer multi-layer perceptron (MLP) is applied in the feedforward step and is also enhanced by an element-wise gate. The edge feature is then updated with the element-wise production of $\mathbf{q}$ and $\mathbf{k}$ (Fig. \ref{fig:coorupdate}\textbf{d}).

\subsubsection{Coordinate Update Block}\label{sec4.3.3}

The coordinate update block updates the coordinates of the nodes (atoms) in 3-D space based on the attentions derived from the feature update block, as also introduced in \cite{EGNN} (Fig. \ref{fig:coorupdate}\textbf{e-f}). 
To be specific, the attention is transformed to a one-dimensional distance gradient, indicating the change in distance between the central node and each of its neighboring nodes. Then, the coordinate of a certain central node is updated by the sum of these distance gradients calculated with all neighboring nodes, as written by,

\begin{equation} \label{eq1}
\mathbf{\Delta}_h = \underset{j\in Neighbor(i)}{\sum}\frac{\mathbf{x}_i-\mathbf{x}_j}{\Vert \mathbf{x}_i-\mathbf{x}_j\Vert_2}\Phi(\mathbf{q}_{ij} \odot \mathbf{k}_{ij}), 
\end{equation}

\begin{equation} \label{eq2}
\mathbf{x}_i^{'} = \mathbf{x}_i + \sum_{h=1,...,N_h}{\mathbf{\lambda}_h\mathbf{\Delta}_h},
\end{equation}

where $\mathbf{x}_i$ denote the coordinate of a central node $i$, $\mathbf{x}_j$ denote the coordinate of a neighbouring node $j$,
$\mathbf{q}_{ij}$ and $\mathbf{k}_{ij}$ are the Query and Key for $i$ and $j$, respectively, as obtained in the feature update block. The element-wise multiplication ($\odot$) of $\mathbf{q}_{ij}$ and $\mathbf{k}_{ij}$ are then transformed to a single distance variable by a linear layer $\Phi$, and multiplied by the direction of a central node's coordinate ($\mathbf{x}_i$) to one of its neighboring node's coordinate ($\mathbf{x}_j$). The coordinate update $\mathbf{\Delta}_h$ for node $i$ is then obtained by aggregating $\mathbf{\Delta}$ for all $i$'s neighbouring nodes ($Neighbor(i)$), where $i$ is not equal to $j$. $\mathbf{x}_i$ is then updated by the weighted sum of all $\mathbf{\Delta}_h$ to $\mathbf{x}_i^{'}$. Finally, the coordinates were updated using MHA with $\mathbf{\lambda}_h$ weighted $N_h$ heads.

\subsection{Correlation-enhanced graph learning}\label{semi_maintext}
We introduce a novel training paradigm, namely correlation-enhanced biomolecular graph learning, that simultaneously learns both the graph features and the 3D structures for proteins and SOMs. 
In brief, in each training iteration, the network is trained with a half-to-half chance by a labeled sample (\ie, a native complex) or an unlabeled sample (\ie, a randomly paired protein and ligand). We use the Monte Carlo method to choose one cycle to optimize the network parameters, as also adopted by \cite{AF2}. Details of the training schedule can be found in Suppl. Sec.\ref{suppl_train}

\subsubsection{Training with native complexes}

Given a ligand containing $N_{lig}$ atoms (nodes). In each iteration, an index is created for mapping the nodes from the predicted pose to the native pose, ensuring the loss is calculated on atoms with the same chemical identity. Then, the loss for structure prediction is calculated by averaging the differential between the predicted and native ligand coordinates using the atom mapping (denote as $\mathcal{L}_{coor}$), 

\begin{equation}
\mathbf{\mathcal{L}}_{coor} = \frac{1}{N_{lig}}\underset{i \in {Nodes(lig)}}{\sum}\Vert \mathbf{x}_i^{pred}-\mathbf{x}_i^{true}\Vert_2.\\
\end{equation}

where $Nodes(lig)$ denotes the set of all ligand nodes. However, many ligands have symmetric structures. In such cases, multiple equivalent indexes can be created for one ligand (see an example in Suppl. Fig. \ref{fig:sym}). To solve the problem, we design a specific loss function, \ie, the symmetric-aware loss (denote as $\mathcal{L}_{sym}$), to update the coordinates, which can be formally written as,

\begin{equation} \label{eq4}
\mathbf{\mathcal{L}}_{sym} = \underset{sym}{\min}(\mathbf{\mathcal{L}}_{coor,s}), s\in \{1,...N_s\}.\\
\end{equation}

where $\underset{sym}{\min}(\cdot)$ calculates the losses for $N_s$ equivalent indexes of the ligand and chooses the minimum one. The final loss for one complex is calculated as a weighted sum of the following three terms: (1) an average of $\mathbf{\mathcal{L}}_{sym}$ calculated with the outputs of all coordinate update blocks; (2) a $\mathbf{\mathcal{L}}_{sym}$ calculated with the final coordinate prediction in the chosen cycle; and (3) an affinity loss (see Suppl. Methods for details).

\subsubsection{Training with randomly paired proteins and ligands}

For an unlabeled sample, the loss function (denote as $\mathbf{\mathcal{L}}_{semi}$) consists of two parts, \ie, the Masking-based Complex Modeling (MCM, denote as $\mathbf{\mathcal{L}}_{mask}$) and the Denoising-based Protein structure Reconstruction (DPR, denote as $\mathbf{\mathcal{L}}_{noise}$). Then, $\mathbf{\mathcal{L}}_{self}$ can be formally written as,

\begin{equation} \label{eq_pretrain_loss}
\mathbf{\mathcal{L}}_{self} = \mathbf{\mathcal{L}}_{mask} + \mathbf{\mathcal{L}}_{noise}.
\end{equation}

These losses are designed to prepare the network parameters in two aspects. 
First, the MCM loss $\mathbf{\mathcal{L}}_{mask}$ is proposed to guide the model to recognize the atom properties. A portion of nodes and edges are masked with a specific token, that encourages the model to predict the original classes of the nodes and edges, as shown in Fig. \ref{fig:semi}\textbf{a}. 
Next, the DPR loss $\mathbf{\mathcal{L}}_{noise}$ aims to learn representative features for the experimentally solved protein structures, in which random spatial noises are applied to part of the protein nodes, that encourage the model to predict the original position of these nodes, as shown in Fig. \ref{fig:semi}\textbf{b}. 

\subsubsection{Training \& evaluation for different tasks}\label{sec_main_text_vs}

\noindent\textbf{CASF-2016.} For the screening power, the model starts with the model evaluated on the core set for the structure prediction task, where an additional task is included, \ie, predicting the probability of the SOM to be a true ligand of a protein, to be complementary to the structure prediction task. The final screen score is obtained by jointly considering this probability and affinity, as the product of these two predictions. With the same training set for the structure prediction task, all proteins and SOMs in PDBbind are paired as positive or negative samples. Proteins with the same UniProt IDs or the same protein names share the same positive SOMs, and SOMs with the same 3-letter PDB IDs share the same positive proteins. The rest pairs are labeled as negative samples. (see Suppl. Methods for details). For the scoring power, we report Pearson R for affinity estimations. For the docking power, we use the \ourmodelscore to identify the near-native pose in the CASF-2016 decoy set.

\noindent\textbf{Structure prediction and virtual screening for \mpronoblank.}
We use five-fold cross-validation for structure prediction and virtual screening of \mpronoblank. The training details of \ourmodel
can be found in Suppl. Methods.

\subsection{Metrics}\label{Metrics}
The Root Mean Square Deviation (RMSD) value between the predicted and native structures is used to evaluate the structure predictions. The RMSD value of $2$\AA~is widely adopted in related works\cite{wang2016comprehensive,su2018comparative} as a standard to determine the success or failure of the prediction, as also adopted in this work. 

The screening power was evaluated on $57$ proteins coupled with $285$ SOMs in CASF-$2016$. In this dataset, each protein has $5$ ligands as true binding cases, then, the rest of the ligands are regarded as negative ones. The performance was assessed by the percentage of best ligands included in $1\%$, $5\%$, and $10\%$ of the top-ranked ligands\cite{su2018comparative}, and the best ligand is the ligand of the highest affinity among $5$ true ligands for each protein. 
Besides, the enhancement factor (EF) is also used to assess the ability of screening which was introduced in Ref.\cite{su2018comparative}. In brief, EF is used to assess the ability to rank the true ligands on the top ranking position which also uses $1\%$, $5\%$, and $10\%$ as cutoffs.

\subsection{Configurations of docking tools}\label{docking_config}
For docking tools, the ligands were firstly rotated $180^{\circ}$ through the Z-axis with their original conformation\cite{wang2016comprehensive}, followed by energy minimization with Open Bable toolbox\cite{obabel}. The docking sites were determined by the native positions of ligands. The amount of generated poses for a certain ligand was set to $20$, as also adopted in other works\cite{lim2019predicting,wang2016comprehensive,morrone2020combining}. The number of CPU cores was set to $1$ to calculate the time cost for prediction (\ie, inference time). The outputs were analyzed using RDKit\cite{rdkit}. Settings details of all docking tools are introduced in Suppl. Sec. \ref{docking_config_suppl}.





\subsection{Data availability}
PDBbind (version $2020$) and CASF-2016 are available at http://www.pdbbind.org.cn. The PDBbind-CrossDocked-Core set is available at https://github.com/sc8668/ml\_pose\_prediction. The \mpro complex structure benchmark is available at https://github.com/shanizev/Benchmarking-SARS-CoV-2. Experimentally measured enzymatic activity of \mpro is available at NCATS\cite{kc2021machine} (https://opendata.ncats.nih.gov/covid19/assays).

\subsection{Code availability}
The source code of \ourmodel is available under an open-source license at https://gitfront.io/r/LigPose/kMWuV4DW6JpE/LigPose4Review/.

\bmhead{Supplementary information}

This paper accompanies supplementary information in the "Supplementary Material".

\bmhead{Acknowledgments}

This work was supported in part by the National Nature Science Foundation of China under Grant No. 62106101. This work was also supported in part by the Natural Science Foundation of Jiangsu Province under Grant No. BK20210180.

\bmhead{Author contributions}
K.H. and J.Z. led the research. K.H. and T.D. contributed to the idea and developed the method. T.D. developed data processing and analytics. K.H., T.D., J.W., and J.Z. wrote the paper.

\bmhead{Correspondence and requests for materials} should be addressed to Junfeng Zhang and Jinhui Wu.

\bmhead{Competing interests}
The authors declare no competing interests.

\clearpage

\begin{appendices}


\section{Supplementary Materials}\label{sec5}

\subsection{Notations}\label{sec5.0}

We summarize the notations present in this paper in Table \ref{variable}.

\subsection{Supplementary Methods}\label{sec5.1}

\ourmodel consists of three key components (see main text Fig. \ref{fig:framework}\textbf{d}), 1) a sampling and recycling strategy, 2) a stochastic coordinate initialization combined with a symmetric-aware loss function, and 3) a multi-task training scheme including the self-supervised learning on large-scale unlabeled data. The implementation details of these components will be introduced below.

\subsubsection{Sampling and recycling}\label{sec5.1.1}

To reduce the computational burden of handling a large number of protein atoms, we proposed a sampling strategy for protein atoms. 
Specifically, for a given graph representing a protein-ligand pair, a sub-graph including all ligand atoms, and the C$\alpha$ and C$\beta$ atoms derived from the protein pocket, is used to feed the model. These atoms are denoted as \emph{core atoms}. Then, a set of atoms that are randomly sampled from the rest of the protein pocket atoms (denote as \emph{context atoms}), are also included in the graph. 
The total number of core and context atoms are denoted as $N_{total}$. All edges are assigned according to the real inter-atom relationship. 
Since the atoms of the complex are not fully used, which may weaken the representative ability of the method, we introduce a recycling strategy to compensate for it. In this case, the method is processed cycles for one complex, and within each cycle, the features and the coordinates of the complex are updated $N_l$ times using the stacked feature and coordinate update blocks.
Besides, in each cycle except the first cycle, the last updated features of the core atoms will be added to a newly sampled sub-graph with the same $\Psi$ as defined in the feature update block, while the last updated features of the context atoms are ignored, to conduct a "partial update". The edge features are updated with the same principle. And the coordinates of the ligand nodes in the newly sampled sub-graph are directly assigned with their last updated values. Then, the sub-graph will be fed to the model again. The recycling will be performed $N_c$ times ($N_c=4$ in this work).

\subsubsection{Feature update block}\label{sec5.1.2}

The feature update block extracts features of all nodes with an architecture based on Graph Transformer\cite{graphtransformer,shi2020masked}. Given an input graph $G = (V,E)$, the feature of the $i$th node $\mathbf{f}_i\in V$, and the feature of the edge between the $i$th node and its $j$th neighbouring node (denote as $\mathbf{e}_{ij}\in E$) are initialized as $\mathbf{f}_i^{init}$ and $\mathbf{e}_{ij}^{init}$, respectively, according to Table \ref{table_feature}. Then, each feature is fed to a linear layer with respect to its feature type (\ie, the protein node feature, the ligand node feature, and the edge feature). The output features of the nodes (denoted as $\mathbf{f}_i^0$) are in the same size $d_f$. The output features of the edges are denoted as $\mathbf{e}_{ij}^0$ with the size of $d_e$.


The features are then processed by the number of $N_l=6$ feature-and-coordinate update blocks. The feature update block conducts two operations, \ie, message aggregation and feedforward. For a message aggregation step in the $l \in \{1,...,N_l\}$th block, the central nodes aggregate the information from the neighboring nodes using both their features and coordinates. In particular, we calculate the Query (denote as $\mathbf{q}$), Key (denote as $\mathbf{k}$) and Value (denote as $\mathbf{v}$) in the block, as written by,
\begin{equation} \label{eq_encode_dist}
\mathbf{d}_{ij}^{l} = \text{Concat}(\text{RBF}(\Vert \mathbf{x}_i^{l-1}-\mathbf{x}_j^{l-1}\Vert_2), \mathbf{e}_{ij}^{l-1}),
\end{equation}

\begin{equation}
\mathbf{k}_{j}^{h,l} = \mathbf{W}_k^{h,l}\mathbf{f}_j^{l-1} \odot \text{LeakyReLU}(\mathbf{W}_{e}^{h,l}\mathbf{d}_{ij}^{l}),
\end{equation}

\begin{equation}
\mathbf{q}_{i}^{h,l} = \mathbf{W}_q^{h,l}\mathbf{f}_i^{l-1},
\end{equation}

\begin{equation}
\mathbf{v}_{j}^{h,l} = \mathbf{W}_v^{h,l}\mathbf{f}_j^{l-1},
\end{equation}

where $\mathbf{x}_i^{l-1}$ and $\mathbf{x}_j^{l-1}$ denote the output coordinates of the $i$th central node and its $j$th neighboring node of the $l-1$th block, respectively. In the rest of Supplementary Materials, $\mathbf{W}$ denote the learnable parameters of Linear layers, and their biases are omitted for simplicity, where $\mathbf{W}_{e}^{l}\in \mathbb{R}^{d_h\times (d_e+d_r)}$, $\mathbf{W}_q^{h,l}, \mathbf{W}_k^{h,l}, \mathbf{W}_v^{h,l}\in \mathbb{R}^{d_h\times d_f}$. Concat$(\cdot)$ denotes the concatenation operation. The Radial Basis Function (RBF) is used to encode the spatial information of coordinates, resulting in $d_r$ dimensions. $d_h$ denote the dimension of the $h$th ($h \in \{1,...,N_h\}$) attention head. In this work, $d_h=d_f/N_h$. $\odot$ stands for element-wise multiplication.

The output of the message aggregation is then calculated by the following steps,
\begin{equation} \label{eqS7}
\mathbf{a}_{ij}^{h,l} = \mathbf{q}_{i}^{h,l} \odot \mathbf{k}_{j}^{h,l}, 
\end{equation}
\begin{equation} \label{eqS8}
\mathbf{\omega}_{ij}^{h,l} = \text{softmax}_j(\frac{1}{\sqrt{d_h}} \mathbf{a}_{ij}^{h,l}\mathbf{1}),  
\end{equation}
\begin{equation} \label{eqS9}
\hat{\mathbf{f}}_i^l = \mathbf{W}_{f,o}^l\underset{h\in1,...,N_h}{\text{Concat}}(\underset{j\in Neighbor(i)}{\sum}\mathbf{\omega}_{ij}^{h,l}\text{Concat}(\mathbf{v}_{i}^{h,l},\mathbf{v}_{j}^{h,l})),
\end{equation}
\begin{equation} \label{eqS10}
\hat{\mathbf{e}}_{ij}^l =  \mathbf{W}_{e,o}^l\underset{h\in1,...,N_h}{\text{Concat}}(\mathbf{\omega}_{ij}^{h,l}\text{LeakyReLU}(\mathbf{W}_{t}^{h,l}\mathbf{d}_{ij}^{l})),
\end{equation}

where $\mathbf{W}_{f,o}^{h,l}\in \mathbb{R}^{d_f\times 2d_f}$, $\mathbf{W}_{e,o}^{h,l}\in \mathbb{R}^{d_e\times d_f}$, $\mathbf{W}_{t}^{l}\in \mathbb{R}^{d_h\times (d_e+d_r)}$. $Neighbor(i)$ denote the set of neighboring nodes of the $i$th central node.

Before feeding the node and edge features to the feedforward step in $l$th block, we apply a novel proposed element-wise gate $\Psi(\cdot)$ on them, as inspired by \cite{shi2020masked}. The gated output of the node feature $\hat{\hat{\mathbf{f}}}_i^l$ can be written as,

\begin{equation} \label{eqS11}
\mathbf{g}^l = \text{sigmoid}(\mathbf{W}_g^l\text{Concat}(\hat{\mathbf{f}}_i^l, \mathbf{f}_i^{l-1}, \hat{\mathbf{f}}_i^l-\mathbf{f}_i^{l-1})),
\end{equation}
\begin{equation} \label{eqS12}
\hat{\hat{\mathbf{f}}}_i^l = \Psi_{f,1}^l(\hat{\mathbf{f}}_i^l, \mathbf{f}_i^{l-1}) = \text{Norm}(\mathbf{g}^l\odot \hat{\mathbf{f}}_i^l + \mathbf{f}_i^{l-1}),
\end{equation}

where $\mathbf{W}_g\in \mathbb{R}^{d_f\times 3d_f}$, Norm$(\cdot)$ denote the Layer Normalization\cite{layernorm}, and $\mathbf{f}_i^l$ is the final updated features in the $l$th block.
$\hat{\hat{\mathbf{f}}}_i^l$ is then fed to the feedforward step, as calculated by,

\begin{equation} \label{eqS14}
\mathbf{f}_i^l = \Psi_{f,2}^l(\mathbf{W}_{f,2}^l\text{LeakyReLU}(\mathbf{W}_{f,1}^l\hat{\hat{\mathbf{f}}}_i^l), \hat{\hat{\mathbf{f}}}_i^l)
\end{equation}
where $\mathbf{W}_{f,1}^l$, $\mathbf{W}_{f,2}^l\in \mathbb{R}^{d_f\times d_f}$.
The edge feature $\mathbf{e}_{ij}^{l}$ is calculated similarly.



\subsubsection{Coordinate update block}

The coordinate update block updates the coordinate of the nodes (atoms) in 3-D space along with the feature update block, as inspired by \cite{EGNN}. Specifically, for the $l$th block, the coordinates are updated as follows,

\begin{equation} \label{eqS17}
\mathbf{\Delta}_h^l = \underset{j\in Neighbor(i)}{\sum}\frac{\mathbf{x}_i^{l-1}-\mathbf{x}_j^{l-1}}{\Vert \mathbf{x}_i^{l-1}-\mathbf{x}_j^{l-1}\Vert_2}\mathbf{W}_x^l\mathbf{a}_{ij}^{h,l},
\end{equation}
\begin{equation} \label{eqS18}
\mathbf{x}_i^l = \mathbf{x}_i^{l-1} + \sum_{h\in1,...,N_h}{\mathbf{\lambda}_h^l\mathbf{\Delta}_h^l},
\end{equation}

where $\mathbf{a}_{ij}^{h,l}$ is calculated in the feature update block (refer to Eq. \ref{eqS7}), $\mathbf{x}$ denote the coordinate, $\mathbf{W}_x^l\in \mathbb{R}^{1\times d_h}$. Note that only the coordinates of the ligand nodes are updated.

\subsubsection{Affinity prediction and virtual screening}\label{aff_scr_layer}

\ourmodel maintains a multi-task learning framework to simultaneously predict the structure, affinity score, and binding power. Thus, it also estimates the affinity of the protein-ligand complex as the auxiliary task, to be complementary to the coordinate prediction. In this case, the pooled node features $\mathbf{f}_i^{N_l}$ and edge features $\mathbf{e}_j^{N_l}$ are fed to a two-layer multi-layer perceptron, as calculated by,

\begin{equation} \label{eq_pooling}
\mathbf{r} = \text{Norm}(\text{Concat}(\frac{1}{N_f}\underset{i\in 1,...,N_f}{\sum}\mathbf{f}_i^{N_l},\frac{1}{N_e}\underset{j\in 1,...,N_e}{\sum}\mathbf{e}_j^{N_l}))
\end{equation}

\begin{equation} \label{eqS19}
\mathbf{y}_{aff} = \text{ReLU}(\mathbf{W}_{a,2}\text{LeakyReLU}(\mathbf{W}_{a,1}\mathbf{r})),
\end{equation}
where $\mathbf{y}_{aff}$ denote the prediction of the affinity, $\mathbf{W}_{a,1}\in \mathbb{R}^{(d_f+d_e)\times (d_f+d_e)}$, $\mathbf{W}_{a,2}\in \mathbb{R}^{1\times (d_f+d_e)}$, and $N_f$, $N_e$ denotes the number of nodes and edges, respectively. 

To screen the possible drugs, we provide a probability estimation (denoted as $\mathbf{y}_{bind}$) to indicate whether a SOM binds to a protein. The probability estimation also served as an additional task, similar to the affinity estimation task, with the ReLU layer replaced by the Sigmoid layer.

\subsection{Training}
\label{suppl_train}

\subsubsection{Correlation-enhanced graph learning}\label{suppl_semi}

\paragraph{Training with native complexes}

Two losses are used in the training stage, \ie, the symmetric-aware loss for structure prediction (as introduced in \ref{preprocess}), and the affinity loss for binding affinity prediction.
The affinities are collected by PDBbind, including $Ki$, $Kd$, and $IC50$. Their negative log scale is used for training and evaluation. Then, the affinity loss is predicted for each protein-ligand pair, as defined by,

\begin{equation} \label{eqS22}
\mathbf{\mathcal{L}}_{aff} = \Vert \mathbf{y}_{aff}^{pred}-\mathbf{y}_{aff}^{true}) \Vert^2_2.
\end{equation}

Then, the final loss of the network is defined as,

\begin{equation} \label{eqS23}
\mathbf{\mathcal{L}} = \gamma_1(\frac{1}{N_l-2}\underset{l\in {2,...,N_l-1}}{\sum}\mathbf{\mathcal{L}}_{sym}^l+\mathbf{\mathcal{L}}_{sym}^{N_l}), +\gamma_2\mathbf{\mathcal{L}}_{aff},
\end{equation}

where $\mathbf{\mathcal{L}}_{sym}$ is the proposed symmetric-aware loss calculated with coordinate output with the new $N_{l}$ blocks as defined in the main text (refer to Main Text Eq. \ref{eq4}) and the equivalent indexes used for $\mathbf{\mathcal{L}}_{sym}$ are obtained from RDKit\cite{rdkit}.
$\gamma_1$ and $\gamma_2$ are weight parameters to balance the losses. 

We adopt a two-step training schedule. The first step limits the maximum number of nodes to $200$, then the second step increases this number. The full training schedule is listed in Table \ref{training_schedule_pdbbind}.

$2200$ random samples in the training set (when testing on the core set) are used to tune the hyper-parameters of the backbone of \ourmodellight without training with unlabeled data. The search space is shown in Table \ref{hyperparameter_search_space}. For \ourmodelnoblank, we simply increase the complexity of it on hidden size, number of cycling, and number of max nodes. The chosen architectures are shown in Table \ref{hyperparameters_architecture}. We used the Adam optimizer with an exponentially decayed learning rate, with a decay rate of $0.99$. Besides, all coordinates and affinities are rescaled to $1/10$ of their original values for stable training. 
And these hyper-parameters are preserved in all downstream tasks.

\paragraph{Training with randomly paired proteins and ligands}

Learning on the unlabeled data aims to encourage the model to learn graph features and biomolecular atom correlations simultaneously via self-supervised learning. The entire self-supervised learning loss function (denote as $\mathbf{\mathcal{L}}_{self}$) consists of two parts, \ie, the Masking-based Complex Modeling (MCM, denote as $\mathbf{\mathcal{L}}_{mask}$) and the Denoising-based Protein structure Reconstruction (DPR, denote as $\mathbf{\mathcal{L}}_{noise}$). Also, to further generalize the model to very strict conditions of non-similar seen molecules, we pre-trained \ourmodel with docking-based structures (see Sec. \ref{Sec:docking-based}).

\noindent\textbf{Masking-based complex modeling (MCM).}
In a molecular graph, the attribute of a node can be inferred from the contextual nodes and edges\cite{hu2019strategies}. 
Therefore, we train the network by masking part of the nodes and edges in a protein-ligand pair with a specific token [MASK], \ie, the original features are zeros after masked. The masking ratio is $15\%$ for both nodes and edges. This MCM stage encourages the network to learn distinct atom features for the proteins and SOMs. The output features of the node $\mathbf{f}_i^{N_l}$ and the edge $\mathbf{e}_j^{N_l}$ followed by a two-layer MLP are then used to predict their true identical attributes, \ie, the element types of SOM nodes (denote as $\mathbf{\mathcal{L}}^{l\_elem\_type}_{mask}$), the atom and residue types of protein nodes (denote as $\mathbf{\mathcal{L}}^{p\_atom\_type}_{mask}$ and $\mathbf{\mathcal{L}}^{p\_res\_type}_{mask}$). 
For the edges, the network is trained to predict their bond types (denote as $\mathbf{\mathcal{L}}^{bond\_type}_{mask}$). The protein-SOM edges are excluded in this MFP stage. Notably, masking is performed in both directions for edges, \ie, the two edges connected same nodes are masked together. If an atom type involves symmetry, only one of them will be preserved (\eg, $CG1$ and $CG2$ atoms in valine). Additionally, the distance features involved in masking will not be masked.
At least one protein node, one SOM node, two protein edges, and two SOM edges are masked in each iteration. We use the Focal loss\cite{lin2017focal} to calculate the loss.
The final $\mathbf{\mathcal{L}}_{mask}$ are weighted as:
\begin{equation} \label{eq_masking_loss}
\mathbf{\mathcal{L}}_{mask} = \mathbf{\mathcal{L}}^{p\_atom\_type}_{mask} + \mathbf{\mathcal{L}}^{p\_res\_type}_{mask} + \mathbf{\mathcal{L}}^{l\_elem\_type}_{mask} + \mathbf{\mathcal{L}}^{bond\_type}_{mask}
\end{equation}

\noindent\textbf{Denoising-based protein structure reconstruction (DPR).} 
This strategy encourages the model to reconstruct the original protein structure when spatial noises are applied to the protein nodes. Specifically, a random position noise ($noise \sim N\left(0, \sigma^{2}\right)$, $\sigma=2$\AA) is applied. To calculate the loss, the average distance between the denoised position and the real position is calculated.
In this work, the noisy protein nodes are randomly sampled in each iteration, with a sampling ratio the same as the masking rate.

\subsubsection{Downstream tasks}

\paragraph{Affinity estimation}
We perform post-training of $3$ epochs for \ourmodelnoblank, with higher loss weight for affinity. The training schedule is listed in Table \ref{training_schedule_down}.

\paragraph{Virtual screening on CASF-2016}
We train the model for virtual screening using Focal loss\cite{lin2017focal} with a weight $\gamma_3$ after the training of the structure prediction task. The model predicts the probability of a SOM to be a true ligand for a certain protein, as described in \ref{sec_main_text_vs}. In each training iteration, we balance the positive and negative pairs to be equally presented. Please note that all pairs containing data from the core set are not used for training. Also, $\mathbf{\mathcal{L}}_{sym}$ and $\mathbf{\mathcal{L}}_{aff}$ are used if the native complex structures and affinities can be provided. The specific hyper-parameter setting for virtual screening is shown in Table \ref{training_schedule_down}, while the others are the same as the structure prediction task. The training schedule is listed in Table \ref{training_schedule_down}.

\paragraph{Structure prediction for \mpronoblank-ligand complex}
We train the \ourmodel with similar settings as described in Sec. \ref{suppl_semi} with $11$ structures of SARS-CoV and MERS-CoV collected from PDB. The training schedule is listed in Table \ref{training_schedule_down}.

\paragraph{Virtual screening on \mpro enzymatic activity} 
We perform five-fold cross-validation to the \mpro enzymatic activity data.
The training process is similar to CASF-2016. The \mpro structures randomly paired with SOMs are used as the inputs. The training schedule is listed in Table \ref{training_schedule_down}. For ImageMol, starting with its pre-trained weights, we adopt a set of fine-tuning parameters suggested in ImageMol work\cite{zeng2022accurate}

\subsection{Evaluation}

In evaluation, we use an ensemble strategy for \ourmodelnoblank, \ie, predicting $N_{ens}=10$ times for a given complex.



\subsection{Settings for ablation study}
We perform an ablation study for \ourmodel by evaluating the success rate on the core set. We report the performance after removing the corresponding part in the Table. \ref{tableabalation}.

\subsection{Settings for energy minimization}
We perform a constraint energy minimization with Merck Molecular Force Field\cite{halgren1996merck} in RDKit\cite{rdkit}, with the \emph{maxIters=$20$} and \emph{forceConstant=$100$}.

\subsection{Settings for docking tools} \label{docking_config_suppl}
The basic settings (\eg, number of generated poses) for all docking tools are introduced in Sec.\ref{docking_config}. Here, we describe the detailed settings below.

\noindent\textbf{Smina} All proteins and ligands were converted to \emph{pdbqt} files as describedin \cite{vinaprepare}. The docking sites of the ligands were set by an automatic box creation tool\cite{smina} using their native poses. The rest of the parameters were set by default. 

\noindent\textbf{LeDock} The input files were automatically generated by Lepro\cite{zhang2016enriching}, including adding hydrogens and defining binding pockets. The rest input parameters are set to default.

\noindent\textbf{AutoDock} All proteins and ligands were converted to \emph{pdbqt} files with the same method as Smina. Grid maps were calculated by autogrid4 with default settings. The rest input parameters are set to default.

\noindent\textbf{rDock} We adopted a standard docking protocol introduced in Ref.\cite{wang2016comprehensive} with a system definition file provided in the guidance of rDock.


\subsection{Investigation of the protein-ligand interaction}\label{sec4.6}

Non-covalent interactions between proteins and ligands were recognized by PLIP\cite{plip}, including water bridge, hydrophobic interaction, salt bridge, $\pi$-cation, hydrogen bond, $\pi$-stack, and halogen bond. 
To better assess the non-covalent interactions between the predicted and native poses, we define the interaction reproducibility similar to\cite{paggi2021leveraging}, as written by,
\begin{equation} \label{eq21}
\text{Interaction reproducibility} = \frac{1+ \text{number of shared interactions} }{2+ \text{total number of unique interactions}}.
\end{equation}

The predicted ligands were placed in the original \emph{pdb} file and then processed by PLIP to calculate a score of interaction reproducibility. Specifically, the interactions were calculated at the atom level for ligands and the residue level for proteins.

The visualization of attention weights for a single head was analyzed with the non-covalent interactions derived from true ligand pose, and the weight of each protein residue was calculated by the average attention weight of C$\alpha$ nodes.


\clearpage

\subsection{Supplementary results}

\subsubsection{Using as a scoring function}\label{result_score}

\begin{figure}[htbp]
\centering
\includegraphics[width=\linewidth]{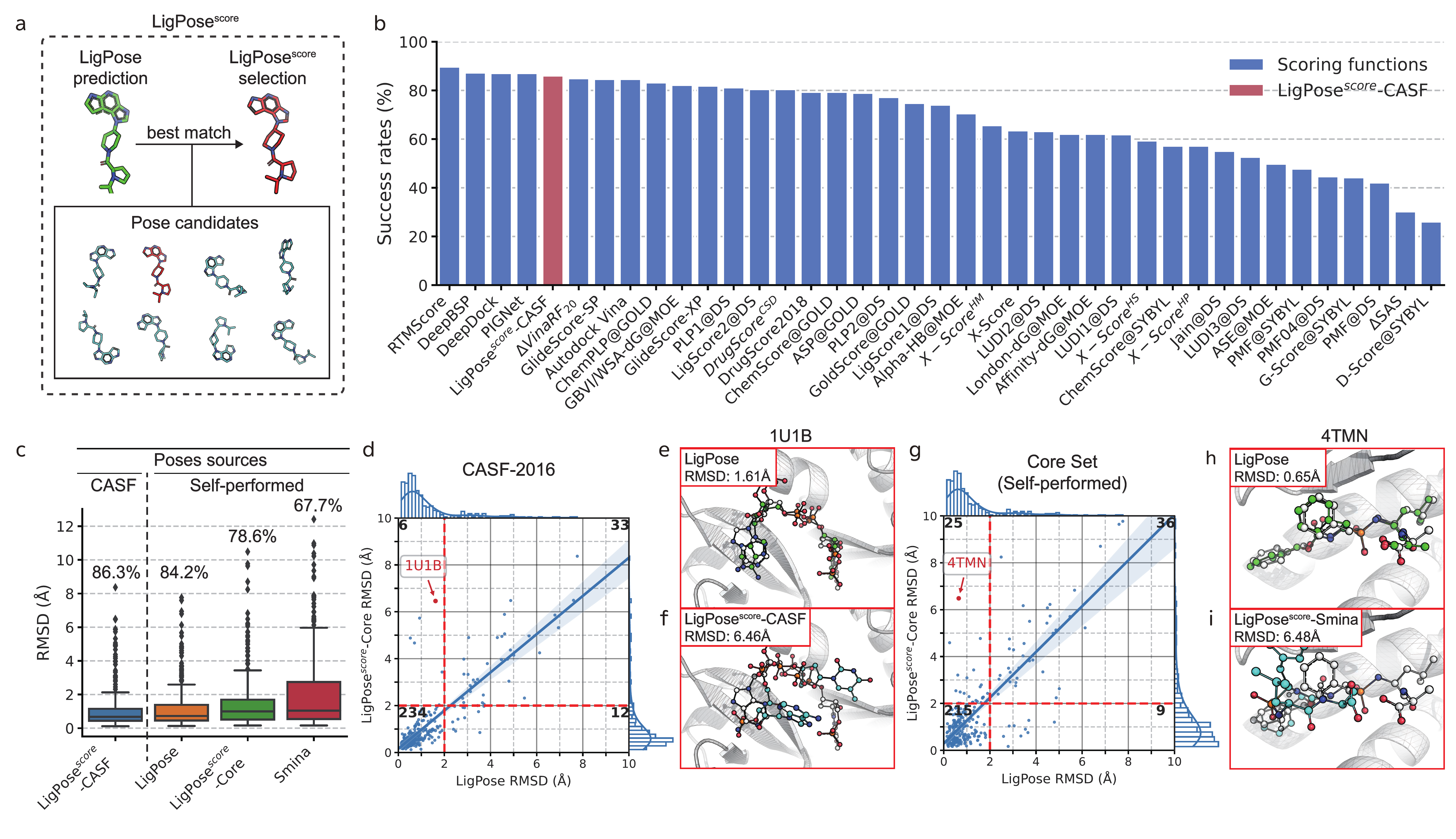}
\caption{\textbf{\ourmodel used as a scoring function.} (\textbf{a}) Brief illustration of \ourmodelscorenoblank, which simply chooses a pose candidate that best matches the prediction of \ourmodel as the top-scored pose. (\textbf{b}) Comparison of success rates between \ourmodelscorenoblank-CASF, and $38$ well-established scoring functions on CASF-2016 benchmark. (\textbf{c}) RMSD distribution and success rate of Smina, \ourmodelnoblank, and \ourmodelscorenoblank-Core, and \ourmodelscorenoblank-CASF. (\textbf{d,g}) Correlation plots of RMSD between predictions of \ourmodel and \ourmodelscore on CASF-2016 (\textbf{d}) and self-performed poses (\textbf{g}), respectively. Each of the figures is partitioned into four regions by two red dashed lines at RMSD of 2\AA, with the sample count marked in the corner. (\textbf{e,f,h,i}) Visualization of two representative samples as pointed out in \textbf{d} and \textbf{g}. Prediction of \ourmodel (\textbf{e,h}), \ourmodelscore (\textbf{f,i}), and the native poses (\textbf{e,f,h,i}) are denoted as green, cyan and white color, respectively. The oxygen and nitrogen atoms are denoted as red and blue color, respectively.}
\label{fig_score}
\end{figure}

Other than directly predicting the poses, we show an extended application of \ourmodelnoblank, \ie, treating it as a scoring function to build \ourmodelscorenoblank, by choosing the pose candidate that best matches the prediction of \ourmodel using the metric of RMSD (see a brief illustration in Fig. \ref{fig_score}\textbf{a}). 
Thus, \ourmodelscore can be composed with docking and evaluated on the Comparative Assessment of Scoring Functions (CASF-2016) benchmark\cite{su2018comparative} derived from the core set. The benchmark is carefully designed to evaluate the capacity of scoring functions, with pose candidates being the clustering centers of docking sampled poses, using RMSD and conformation similarity as the clustering index. 

As suggested by Fig. \ref{fig_score}\textbf{b}, our method (denote as \ourmodelscorenoblank-CASF) with a success rate of $86.3\%$ outperformed $34$ conventional scoring functions (with success rates ranging from $26.0\%$ to $84.9\%$), and was also comparable to the state-of-the-art deep learning-based scoring functions, \ie, RTMScore ($89.7\%$)\cite{RTMScore}, DeepBSP ($87.2\%$)\cite{DeepBSP}, DeepDock ($87.0\%$)\cite{mendez2021geometric}, and PIGNet ($87.0\%$)\cite{moon2022pignet}.
We conclude the slightly lower performance of \ourmodelscore for two reasons. First, \ourmodelscore still needs to generate a near-native pose, instead of explicitly scoring and ranking the pose candidates, which improves the difficulty of the task.
Second, the CASF-2016 benchmark is inadequate to assess the structure prediction ability for real production purposes. Since the RMSD is inaccessible in real-world applications, the native complex structure could not be predetermined during virtual measurement. As previously discussed for Fig. \ref{fig:betterthandocking}\textbf{e}, several works use self-performed poses to partially solve the problem. We adopt this setting and evaluate \ourmodelscore on self-performed poses, by performing Smina on the core set (denote as \ourmodelscorenoblank-Core). The RMSD distribution of Smina, \ourmodelscorenoblank-Core, and \ourmodel are shown in Fig. \ref{fig_score}\textbf{c}, with \ourmodelscorenoblank-CASF as the baseline. As shown in Fig. \ref{fig:betterthandocking}\textbf{d} and Fig. \ref{fig_score}\textbf{c}, without selective poses, the success rates of all methods are decreased compared with those in CASF-2016, ranging from $55.1\%$ to $84.2\%$. \ourmodelscorenoblank-Core showed a greatly improved success rate ($78.6\%$) compared with Smina ($67.7\%$) and was also comparable to the best performing scoring function (\ie, RTMScore ($80.7\%$)\cite{RTMScore}). Moreover, \ourmodel ($84.2\%$) outperformed RTMScore and \ourmodelscorenoblank-Core with notable higher success rates of $3.5\%$ and $5.6\%$, respectively, indicating that the end-to-end learning paradigm of \ourmodel is more effective in real applications without pre-designing the pose candidates. 

Furthermore, we plotted the correlation of RMSD in Fig. \ref{fig_score}, between \ourmodel and \ourmodelscorenoblank-CASF (\textbf{d}) / -Core (\textbf{g}), each with a typical sample visualized by the predicted structures aside (\textbf{e,f,h,i}). We observed from the figure that docking may mislead the model. For some cases that were correctly predicted by \ourmodel (\ie, with RMSD $<2$\AA), \ourmodelscore reshaped them to wrong structures, especially on self-performed poses (with more samples in the upper-left regions in \textbf{g} compared with \textbf{d}). These results also indicate that docking-based sampling and scoring have inevitably ignored limitations for structure prediction in real practice.

\subsubsection{Affinity estimation}\label{result_aff}
\begin{table}[htbp]
  \caption{Performance comparison for affinity estimation on the core set of PDBbind. Methods that require native complex structures were listed in the second set of rows. Methods without using the native complex structures were listed in the last set of rows. The term "Native" indicates the methods that use the native complex structures for affinity estimation, "Docking" indicates they use the docking-sampled complex structures, and "None" indicates they use none of the above complex structures.}
  \label{table_aff}
    \resizebox{\textwidth}{!}{
    \begin{tabular}{lcccccc}
    \toprule
    Models  & MAE   & RMSE  & Pearson $R$  &  \multicolumn{3}{c}{Using Complex Structure} \\
    \cline{5-7}
    &&&& Native & Docked & None \\
    \midrule
    Pafnucy\cite{stepniewska2018development}                & $1.13 $   & $1.42 $   & $0.78 $    & \checkmark         &  \\
    TopologyNet\cite{cang2017topologynet}                   & -         & $1.34 $   & $0.81 $    &  \checkmark  &       &  \\
    FAST\cite{jones2021improved}                            & $1.019$   & $1.308$   & $0.810$    & \checkmark  &           &  \\
    OnionNet\cite{zheng2019onionnet}                        & $0.984$   & $1.278$   & $0.816$    & \checkmark  &          &  \\
    InteractionGraphNet\cite{InteractionGraphNet}                           & $0.940$   & $1.220$   & $0.837$    & \checkmark  &        &  \\
    PointTransformer\cite{wang2022point}                    & $0.91 $   & $1.19 $   & $0.852$    & \checkmark  &       &  \\
    \midrule  
    FAST\cite{jones2021improved}       & $1.498$   & $1.871$   & $0.712$    & &   \checkmark      &  \\
    OnionNet\cite{zheng2019onionnet}   & -         & $1.523$   & $0.755$    & &   \checkmark       &  \\
    InteractionGraphNet\cite{InteractionGraphNet}      & $1.150$   & $1.503$   & $0.757$    & &   \checkmark   &  \\
    APMNet\cite{shen2021cascade}                            & $0.998$   & $1.268$   & $0.815$    &    & & \checkmark \\
    \ourmodelaffinitynoblank                     & $0.908$   & $1.121$   & $0.834$    &  &  & \checkmark\\
    \bottomrule
    \end{tabular}}
\end{table}

To further investigate the scalability of \ourmodelnoblank, we also apply it to the task of protein-ligand affinity estimation (denote as \ourmodelaffinitynoblank), another core function of docking. For disease-related proteins, ligands with strong binding interactions (\ie, affinity) can be selected as drug candidates. \ourmodel is feasible to adapt to this task without leveraging native complex structures, by simply adding several task-specific output layers (see Suppl. Sec. \ref{aff_scr_layer} for details). The comparisons of \ourmodelaffinity with several recent structure-based deep learning methods\cite{jones2021improved,InteractionGraphNet,stepniewska2018development,cang2017topologynet,shen2021cascade,zheng2019onionnet,wang2022point} on affinity estimation are summarized in Table \ref{table_aff}, where \ourmodelaffinity achieved the best performance among the competing methods which do not use native structures, with mean absolute error (MAE) of $0.908$, root mean square error (RMSE) of $1.121$ and Pearson $R$ of $0.834$. Compared to the methods that use native structures, \ourmodelaffinity is also comparable or better than the best-performing method PointTransformer (MAE: $0.91$, RMSE: $1.19$, Pearson $R$: $0.852$), indicating its strong ability to learn native-like atom interactions. 
Therefore, without using native complex structures, \ourmodelaffinity can achieve impressive results that largely outperform other methods.

\subsubsection{Ablation study}
\label{Sec:ablation}

As reported in Table \ref{tableabalation}, the performance of \ourmodel decreases after removing our core designs. The cycling and sampling strategy has the greatest impact on the \ourmodelnoblank, making the success rate drop $16.14\%$. The rest designs (self-supervised learning, ensemble, and symmetric mapping) show drops in a range of $1.40\%$ to $3.16\%$.

\subsubsection{Accuracy for unseen bio-compounds}

We report accuracy after removing ligands in the test set according to Tanimoto similarities\cite{bajusz2015tanimoto} to the training set with different cutoffs. As shown in Table \ref{similarity}, the success rate drops by $2.25\%-5.92\%$ with the similarity cutoff of $0.70-1.00$, which is an acceptable performance for mining unseen compounds. 

\subsubsection{Accuracy for similarity-based filtering}
\label{Sec:docking-based}

Here, we investigate the impact of similarity by applying similarity-based filtering for both protein and ligand. The training set is filtered according to its protein and ligand similarity to the test set. As shown in Table \ref{similarity-re}, \ourmodellight~achieves success rate of $78.60\%$ without similarity filtering (\ie similarity cutoff = 1.0). The success rate drops to $19.30\%$ after filtering all complex structures with protein or ligand similar to the test set (similarity cutoff = $0.4$). 

We proposed the docking-based pre-training strategy to leverage the docking-generated structures as the ground-truths for training. This expands the learned chemical space for the model. Then, the models are trained with a selected training set, and the re-docked structures on the test set. We consider two ways to use re-docked structures, \ie, training with top-scored re-docked structures using, and training with the nearest re-docked structures. As shown in Table \ref{similarity-re}, the success rate of the model increased from $19.30\%$ to $37.89\%$ (nearest) and $56.84\%$ (top-scored), which allowed the model to fit the hardest condition. 

It is quite typical for ML/DL-based method's performance to drop significantly when similarity-based filtering is used. For example, DiffDock also does much worse in such strict evaluation context\cite{diffdock}. The primary concern for this comes from the vast chemical space for the combination of protein and ligand. The current experimentally solved structures are limited. For example, the PDBbind (v2020) data set ($N=19443$) can not afford sufficient samples to generalize the prediction to low-similarity structures.

\subsubsection{Energy minimization}

We perform energy minimization to ensure the predicted structures are physically plausible. The success rate shows a slight drop of $1.40\%$, which indicates that the accuracy is dominantly determined by \ourmodelnoblank, and the energy minimization has little impact on the final predictions.

\subsubsection{Training time}
The product model was trained on $3$ NVIDIA A100 GPU devices for about $9$ days.

\subsection{Supplementary tables}\label{sec5.4}

\begin{table}[htbp]
  \caption{\textbf{Notations.}}
  \label{variable}
    \begin{tabular}{lll}
    \toprule
    Notations    & Descriptions                                &  \\
    \midrule
    $N_l$        & Number of blocks                            &  \\
    $N_{ens}$    & Number of ensemble                          &  \\
    $N_c$        & Number of cycles                            &  \\
    $N_h$        & Number of attention heads                   &  \\
    $N_f$        & Number of nodes                             &  \\
    $N_{lig}$    & Number of ligand nodes                      &  \\
    $N_s$        & Number of ligand equivalent indexes         &  \\
    
    $l$          & Index of blocks                             &  \\
    $h$          & Index of heads                              &  \\
    $d_f$        & Dimension of node feature                   &  \\  
    $d_e$        & Dimension of edge feature                   &  \\
    $d_h$        & Dimension of attention head                 &  \\
    $d_r$        & Dimension of RBF encoded spacial information &  \\
    $\gamma_1$   & Weight for symmetric-aware loss             &  \\
    $\gamma_2$   & Weight for affinity loss                    &  \\
    
    \midrule
    
    $\mathbf{f}$          & Node feature                                &  \\
    $\mathbf{e}$          & Edge feature                                &  \\
    $\mathbf{x}$          & Node coordinate                             &  \\
    $\mathbf{q}$          & Query for attention                         &  \\
    $\mathbf{k}$          & Key for attention                           &  \\
    $\mathbf{v}$          & Value for attention                         &  \\
    $\mathbf{\Delta}$     & Distance gradient                           &  \\
    $\mathbf{\lambda}$    & Weight parameter for multi-head attention in coordinate update  &  \\
    
    $\mathbf{\mathcal{L}}$           & Total loss                          &  \\
    $\mathbf{\mathcal{L}}_{coor}$    & Coordinate loss                     &  \\
    $\mathbf{\mathcal{L}}_{sym}$     & Symmetric-aware loss                &  \\
    $\mathbf{\mathcal{L}}_{aff}$     & Affinity loss                       &  \\
    
    $\mathbf{W}$                     & Weight parameter of Linear layer    &  \\
    $\mathbf{1}$ & Column vector of 1s & \\
    
    \bottomrule
    \end{tabular}
\end{table}

\begin{table}[htbp]
  \centering
  \caption{\textbf{Feature list of nodes and edges.} All features are processed with one-hot encoding except the distances. The initialization of distances is described in the Main text Methods. The feature sizes are denoted in brackets.}
  \label{table_feature}
    \begin{tabular}{p{12em}p{12em}p{10em}}
    \toprule
    \multicolumn{2}{c}{Node features} & \multicolumn{1}{c}{\multirow{2}[4]{*}{Edge features}} \\
\cmidrule{1-2}    \multicolumn{1}{l}{Protein node} & \multicolumn{1}{l}{Ligand node} & \multicolumn{1}{c}{} \\
    \midrule
    element type (4)\newline{}atom degree (5)\newline{}implicit valence (5)\newline{}neighboring H atoms (5)\newline{}atom hybridization (3)\newline{}amino acid type (20)\newline{}atom name (37) & element type (10)\newline{}atom degree (6)\newline{}implicit valence (5)\newline{}neighboring H atoms (5)\newline{}atom hybridization (6)\newline{}formal charge (6)\newline{}ring of size (6)\newline{}aromatic type (1)  & covalent connection (1)\newline{}distance (1)\newline{}covalent bond type (5)\\
    \bottomrule
    \end{tabular}
  \label{tab:addlabel}
\end{table}

\begin{table}[htbp]
  \caption{\textbf{Hyperparameters for \ourmodel architecture.}}
  \label{hyperparameters_architecture}
    \begin{tabular}{lll}
    \toprule
    Hyperparameters                         & \ourmodelnoblank   &  \ourmodellight  \\
    \midrule
    Number of feature update blocks         & $6$                &  $6$  \\  
    Number of coordinate update blocks      & $6$                &  $6$  \\  
    Dimension of node feature ($d_f$)       & $768$              &  $160$\\  
    Dimension of edge feature  ($d_e$)      & $384$              &  $80$ \\  
    Number of attention heads ($N_h$)       & $8$                &  $4$  \\  
    Number of ensemble ($N_{ens}$)          & $10$               &  $10$ \\  
    Number of cycles ($N_c$)                & $4$                &  $3$  \\
    \bottomrule
    \end{tabular}
\end{table}

\begin{table}
\caption{\textbf{Hyperparameter search space.}}
\label{hyperparameter_search_space}
\begin{tabular}{ll} 
\toprule
Hyperparameters & Search space \\ 
\midrule
Learning rate & $\{1\times10^{-3}, 1\times10^{-4}, 1\times10^{-5}\}$ \\
MHA heads & $\{1, 4, 8\}$ \\
Blocks & $\{3, 6\}$ \\
Node/edge hidden size & $\{160/80, 80/40\}$ \\
Cycles & $\{1, 2, 3, 4\}$ \\
\bottomrule
\end{tabular}
\end{table}

\begin{table}[htbp]
\centering
\caption{\textbf{Training schedules for structure prediction with PDBbind dataset.} 
}
\label{training_schedule_pdbbind}
\resizebox{\linewidth}{!}{
\begin{tabular}{lllllll} 
\toprule
Tasks & \multicolumn{3}{c}{Evaluation on core set} & \multicolumn{2}{c}{Evaluation on refined set}\\  
\cline(r){2-4} \cline(r){5-6}
& Stage 1 & Stage 2 & Stage 3 & Stage 1 & Stage 2 \\ 
\midrule
Training with additional unlabeled data & $\textbf{\checkmark}$ & $\textbf{\checkmark}$ & $\textbf{\checkmark}$ & $\mathbf{\times}$ & $\mathbf{\times}$ \\ 
Learning rate & $1\times10^{-4}$ & $1\times10^{-5}$ & $1\times10^{-5}$ & $1\times10^{-4}$ & $1\times10^{-5}$ \\
Weight for symmetric-aware loss ($\gamma_1$) & $1.0$ & $1.0$ & $1.0$ & $1.0$ & $1.0$ \\
Weight for affinity loss ($\gamma_2$) & $1.0$ & $1.0$ & $1.0$ & $1.0$ & $1.0$ \\
Batch size & $15$ & $3$ & $1$ & $12$ & $4$\\
Epochs & $75$ & $5$ & $3$ & $200$ & $3$ \\
Number of max nodes & $200$ & $440$ & $700$ & $500$ & $700$\\
\midrule
Model version & $c1$ & $c2$ & $c3$ & $r1$ & $r2$ \\
Parameters initialized from & Random & $c1$ & $c1$ & Random & $r1$ \\
\bottomrule
\end{tabular}
}
\end{table}

\begin{table}[htbp]
\centering
\caption{\textbf{Training schedules for downstream tasks.} The parameter represented by "-" is not used in the corresponding task due to the lack of available data.}
\label{training_schedule_down}
\resizebox{\linewidth}{!}{
\begin{tabular}{llllll} 
\toprule
Tasks & Affinity estimation & \multicolumn{2}{c}{\makecell{Virtual screening \\ on CASF-2016}} & \makecell{Structure prediction \\ for \mpronoblank} & \makecell{Virtual screening \\ for \mpronoblank} \\  
\cline(r){2-2} \cline(r){3-4} \cline(r){5-5} \cline(r){6-6} 
& Stage 1 & Stage 1 & Stage 2 & Stage 1 & Stage 1 \\ 
\midrule
Training with additional unlabeled data & $\mathbf{\checkmark}$ & $\mathbf{\times}$ & $\mathbf{\times}$ & $\mathbf{\checkmark}$ & $\textbf{\checkmark}$ \\ 
Learning rate & $3\times10^{-6}$ & $1\times10^{-5}$ & $5\times10^{-6}$ & $1\times10^{-5}$ & $1\times10^{-5}$ \\
Learning rate decay & $0.90$ & $0.99$ & $0.99$ & $0.99$ & $0.90$ \\
Weight decay & $0$ & $0$ & $0$ & $0$ & $1\times10^{-3}$ \\
Weight for symmetric-aware loss ($\gamma_1$) & $0.1$ & $0.2$ & $0.2$ & $1.0$ & - \\
Weight for affinity loss ($\gamma_2$) & $10.0$ & $1.0$ & $1.0$ & - & - \\
Weight for virtual screening loss ($\gamma_3$) & - & $1.0$ & $1.0$ & - & $1.0$ \\
Batch size & $3$ & $15$ & $3$ & $1$ & $3$ \\
Epochs & $3$ & $50$ & $30$ & $20$ & $15$ \\
Number of max nodes & $440$ & $200$ & $440$ & $700$ & $430$ \\
\midrule
Model version & $a1$ & $vc1$ & $vc2$ & $m1$ & $vm1$ \\
Parameters initialized from & $c2$ & $c2$ & $vc1$ & $c3$ & $c2$ \\
\bottomrule
\end{tabular}
}
\end{table}


\begin{table}
\caption{\textbf{Quantitative comparison of success rate among \ourmodelnoblank, the docking tools, and the hybrid deep learning methods on the PDBbind-CrossDocked-Core set.} The upper panel includes three docking tools with their derived hybrid machine/deep learning methods. The best-performed methods are marked in bold.}
\label{tablecross}
\begin{tabular}{l|ccc}
\toprule
Methods & \multicolumn{3}{c}{Success rates (\%)} \\
\midrule
& Surflex & Glide & Vina \\
\cline(r){2-2} \cline(r){3-3} \cline(r){4-4} 
Docking & 43.8 & 42.2 & 31.8 \\
Vina & 39.8 & 35.9 & 32.1 \\
X-Score & 41.8 & 36.2 & 32.8 \\
Prime-MM/GBSA & 42.7 & 41.0 & 22.5 \\
NNscore\_XGB\_Re & 47.9 & 37.6 & 28.4 \\
NNscore+Rank\_XGB\_Re & 47.8 & 42.1 & 33.6 \\
ECIF+Vina\_XGB\_Re & 47.5 & 41.4 & 31.2 \\
ECIF+Vina+Rank\_XGB\_Re & 50.1 & 46.0 & 36.1 \\
NNscore\_XGB\_Cross & 52.1 & 44.5 & 35.2 \\
NNscore+Rank\_XGB\_Cross & 51.7 & 44.4 & 35.2 \\
ECIF+Vina\_XGB\_Cross  & 50.4 & 45.3 & 34.8 \\
ECIF+Vina+Rank\_XGB\_Cross  & 52.2 & 47.1 & 35.5 \\
DeepDockM  & 48.0 & 45.5 & 37.4 \\
RTMScore  & 51.9 & 49.3 & 40.9 \\
Docking (top 20) & 66.7 & 63.0 & 51.9 \\
\midrule
\ourmodelnoblank & \multicolumn{3}{c}{\textbf{72.0}} \\
\bottomrule
\end{tabular}
\begin{tablenotes}
\footnotesize
*Performances of other methods were directly copied from Ref.\cite{RTMScore}. The suffix "\_Re" indicates training with re-docked poses, and the suffix "\_Cross" indicates training with cross-docked poses.
\end{tablenotes}
\end{table}

\begin{table}
\caption{\textbf{Abalation study.}}
\label{tableabalation}
\begin{tabular}{ll}
\toprule
Settings & Success rates (\%)\\
\midrule
Baseline &  $84.21$\\
no self-supervised training &  $81.05$\\
no cycling and sampling &  $68.07$\\
no ensemble &  $82.11$\\
no symmetric mapping & $82.81$\\
\bottomrule
\end{tabular}
\end{table}

\begin{table}[htbp]
\caption{\textbf{Success rates for novel bio-compounds, after removing ligands with different similarity cutoffs in the test set.}}
\label{similarity}
\begin{tabular}{ll} 
\toprule
Similarity cutoff & \ourmodel success rates (\%)\\  
\midrule
$1.00$ & $84.21$ \\
$0.95$ & $81.63$ \\ 
$0.90$ & $81.96$ \\ 
$0.80$ & $80.12$ \\ 
$0.70$ & $78.29$ \\ 
$0.60$ & $74.12$ \\ 
$0.50$ & $71.05$ \\ 
$0.40$ & $80.00$ \\ 
\bottomrule
\end{tabular}
\end{table}

\begin{table}[htbp]
\caption{\textbf{Success rates on re-docking task for the core set with similarity-based filtered training set.}}
\label{similarity-re}
\resizebox{\linewidth}{!}{
\begin{tabular}{lll} 
\toprule
Protein/ligand similarity cutoffs & Models & Success rates (\%)\\  
\midrule
$1.0/1.0$ & \ourmodellight~ & $78.60$ \\
$0.4/0.4$ & \ourmodellight~ & $19.30$ \\
$0.4/0.4$ & \ourmodellight~+ trained with nearest re-docked structures & $37.89$ \\
$0.4/0.4$ & \ourmodellight~+ trained with top-scored re-docked structures & $56.84$ \\
\bottomrule
\end{tabular}
}
\end{table}

\subsection{Protein and ligand similarity}\label{simmethod}
The protein similarity is calculated by the sequence identity using the Needleman-Wunsch algorithm\cite{needleman1970general}. The ligand similarity is measured using the Tanimoto similarity calculated by Morgan fingerprints\cite{pattanaik2020molecular}.

 \clearpage

\subsection{Supplementary figures}\label{sec5.5}

\begin{figure}[htbp]
\centering
\includegraphics[width=0.9\linewidth]{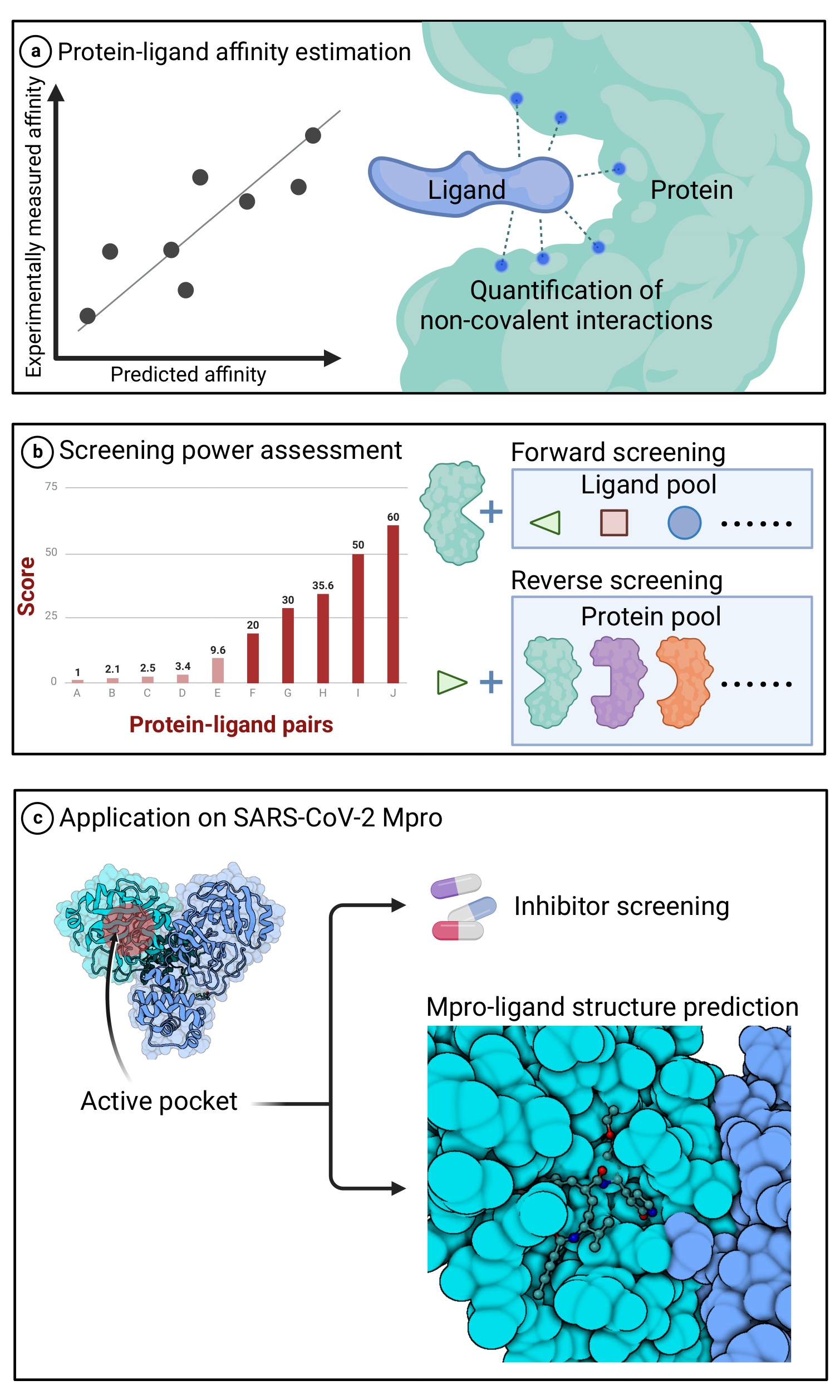}
\caption{\textbf{Multi-task design/evaluation of \ourmodel in drug development based on accurate structure prediction.} (\textbf{a}) Affinity estimation provides strength quantification of protein-ligand interactions. (\textbf{b}) Assessing screening power on forward screening task (\ie, identifying potential binding ligand for a protein) and reverse screening task (\ie, identifying potential binding protein for a SOM). (\textbf{c}) Application of screening and structure prediction for SARS-CoV-2 \mpronoblank.}
\label{fig:multitask}
\end{figure}

\begin{figure}[htbp]
\centering
\includegraphics[width=\linewidth]{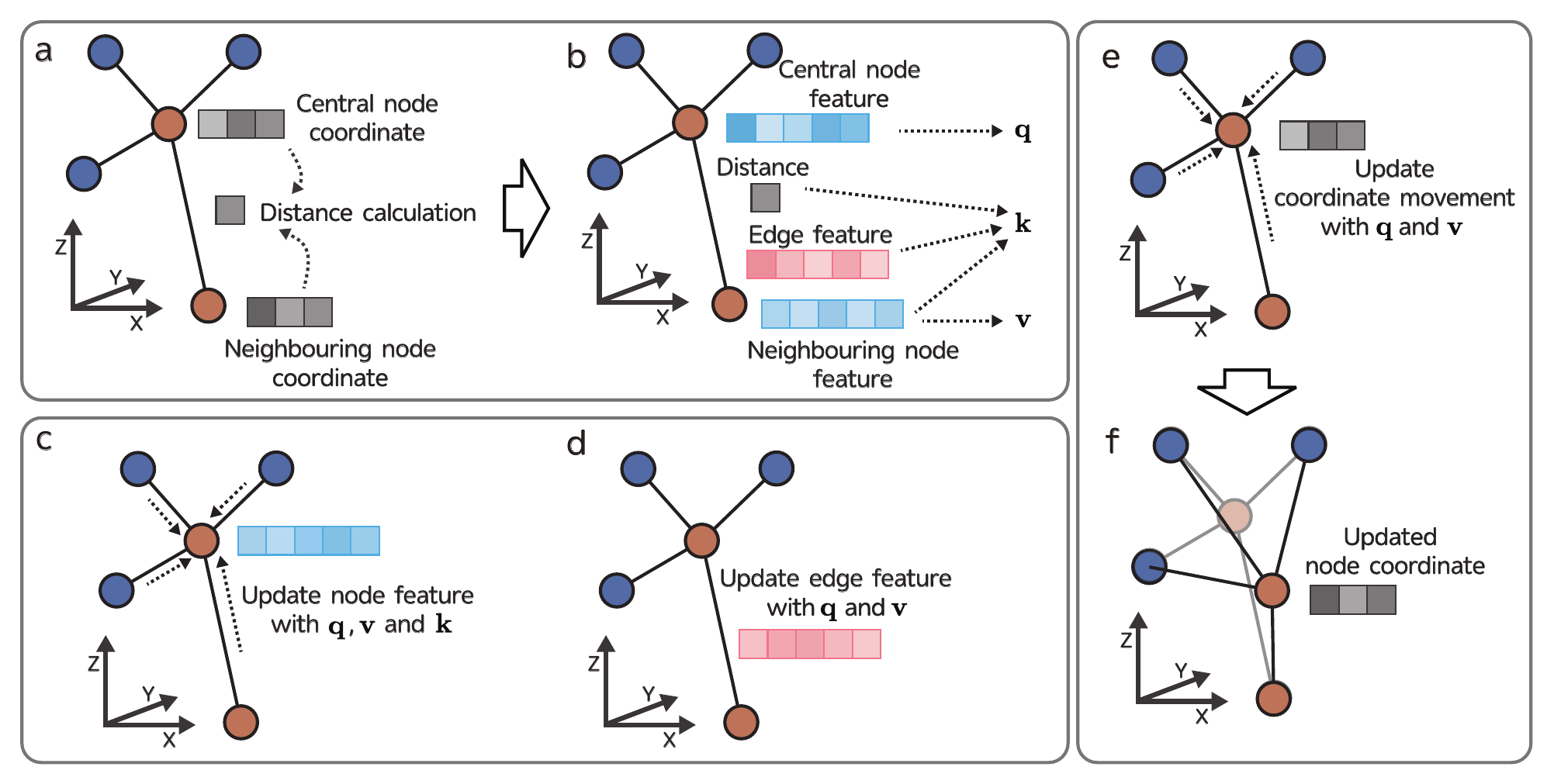}
\caption{\textbf{The updating rules for the nodes and edges of \ourmodelnoblank.} (\textbf{a-f}) Updating features of the central node and its related edges by feature update block.
(\textbf{a}) Calculating pair-wise distances. (\textbf{b}) Calculating attention weights in feature update block. Arrows show the information flow. (\textbf{c}) Updating central node features with message aggregation. (\textbf{d}) Updating edge features. (\textbf{e-f}) Updating coordinates by coordinate update block. (\textbf{e}) Updating central node coordinates. (\textbf{f}) The updated coordinate of the central node.}
\label{fig:coorupdate}
\end{figure}

\begin{figure}[h]
\centering
\includegraphics[width=0.7\linewidth]{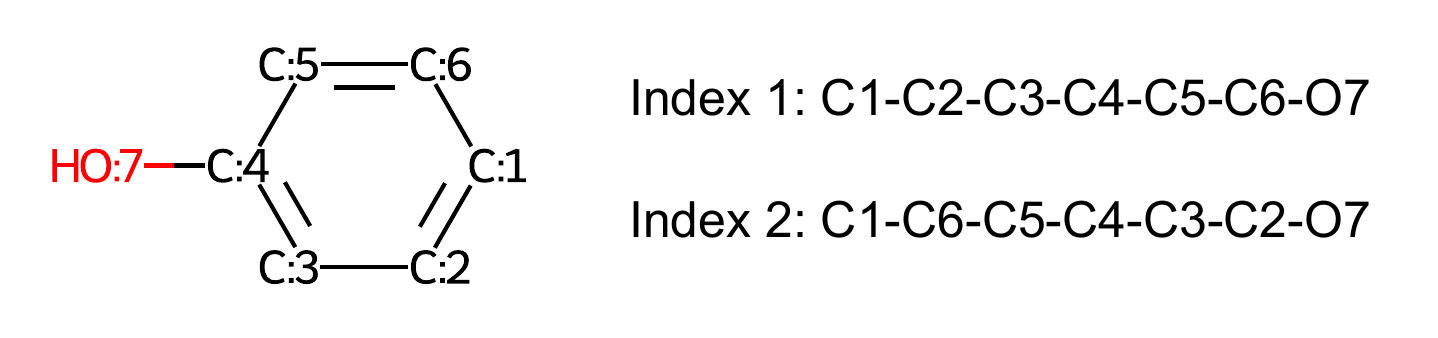}
\caption{\textbf{A symmetric molecule (phenol)}. Two equivalent indexes, \ie, C1-C2-C3-C4-C5-C6-O7 and C1-C6-C5-C4-C3-C2-O7, can be created for this molecule.}
\label{fig:sym}
\end{figure}

\begin{figure}[h]
\centering
\includegraphics[width=\linewidth]{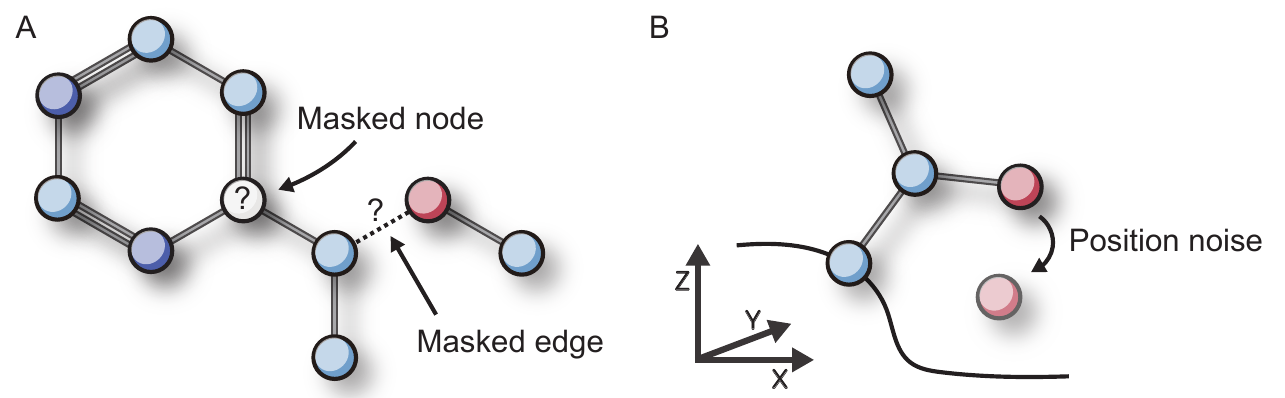}
\caption{\textbf{Schematic of Graph masking and structure denoising of self-supervised learning.} a. Nodes or edges are masked, and the model is encouraged to predict their original attributes (\ie, the atom types and bond types). b. Atoms of pockets are noised in 3D spatial, the model is encouraged to reconstruct their positions.}
\label{fig:semi} 
\end{figure}

\begin{figure}[h]
\centering
\includegraphics[width=\linewidth]{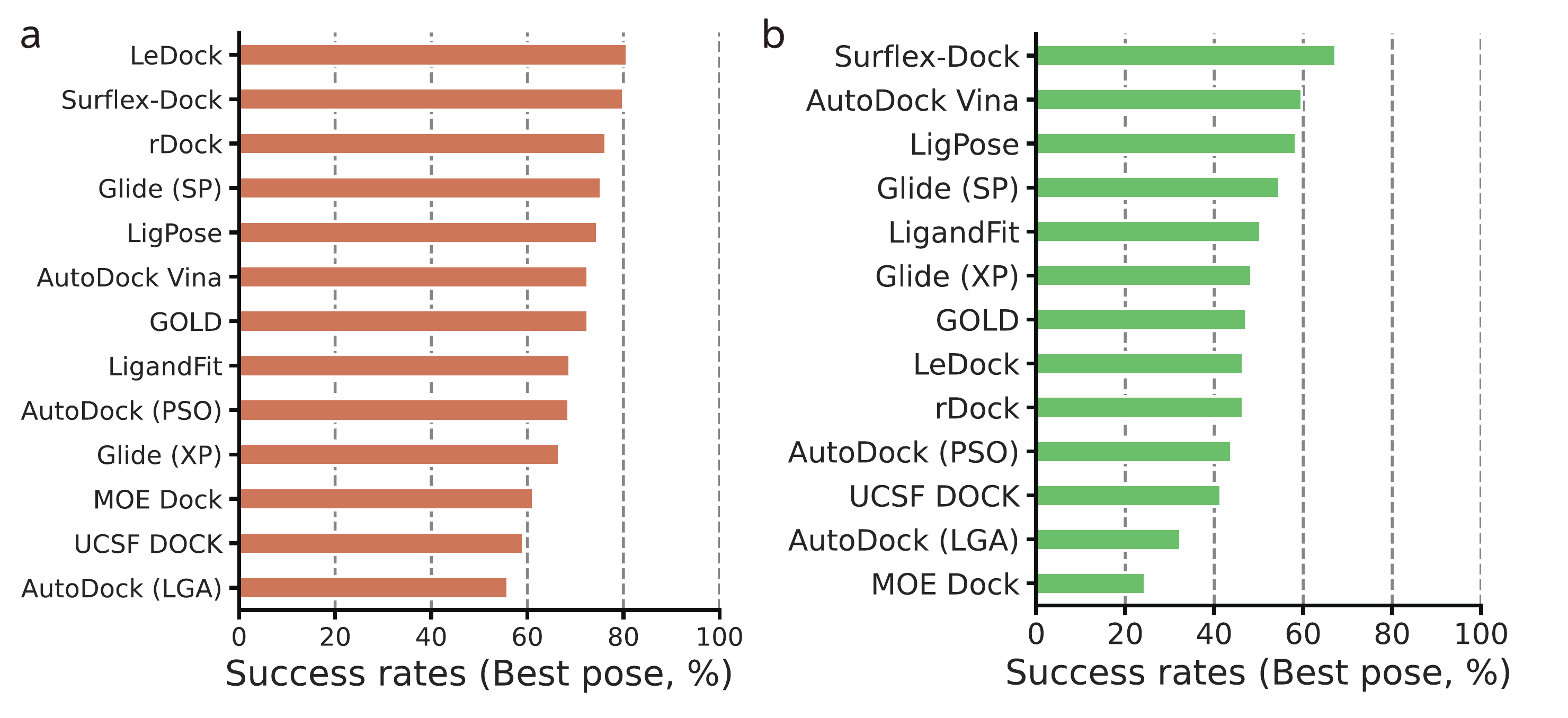}
\caption{\textbf{Quantitative comparison of success rate between \ourmodel and the best pose generated by docking tools on regular organic (a) or peptide/peptide-like (b) ligands.} 
For regular organic ligands, LeDock achieved the highest success rate of $80.8\%$. \ourmodel (success rate of $74.6\%$) ranked the $5$th among all methods.
For peptide/peptide-like ligands, Surflex Dock achieved the highest success rate of $67.3\%$. \ourmodel (success rate of $58.4\%$) ranked the $3$th among all methods. As described in main text Sec. \ref{sec2.2}, these results can be regarded as the upper bound of docking tools, where \ourmodel predicts a greatly higher success rate than $8$ docking tools.}
\label{fig:bestpose} 
\end{figure}

\begin{figure}[h]
\centering
\includegraphics[width=\linewidth]{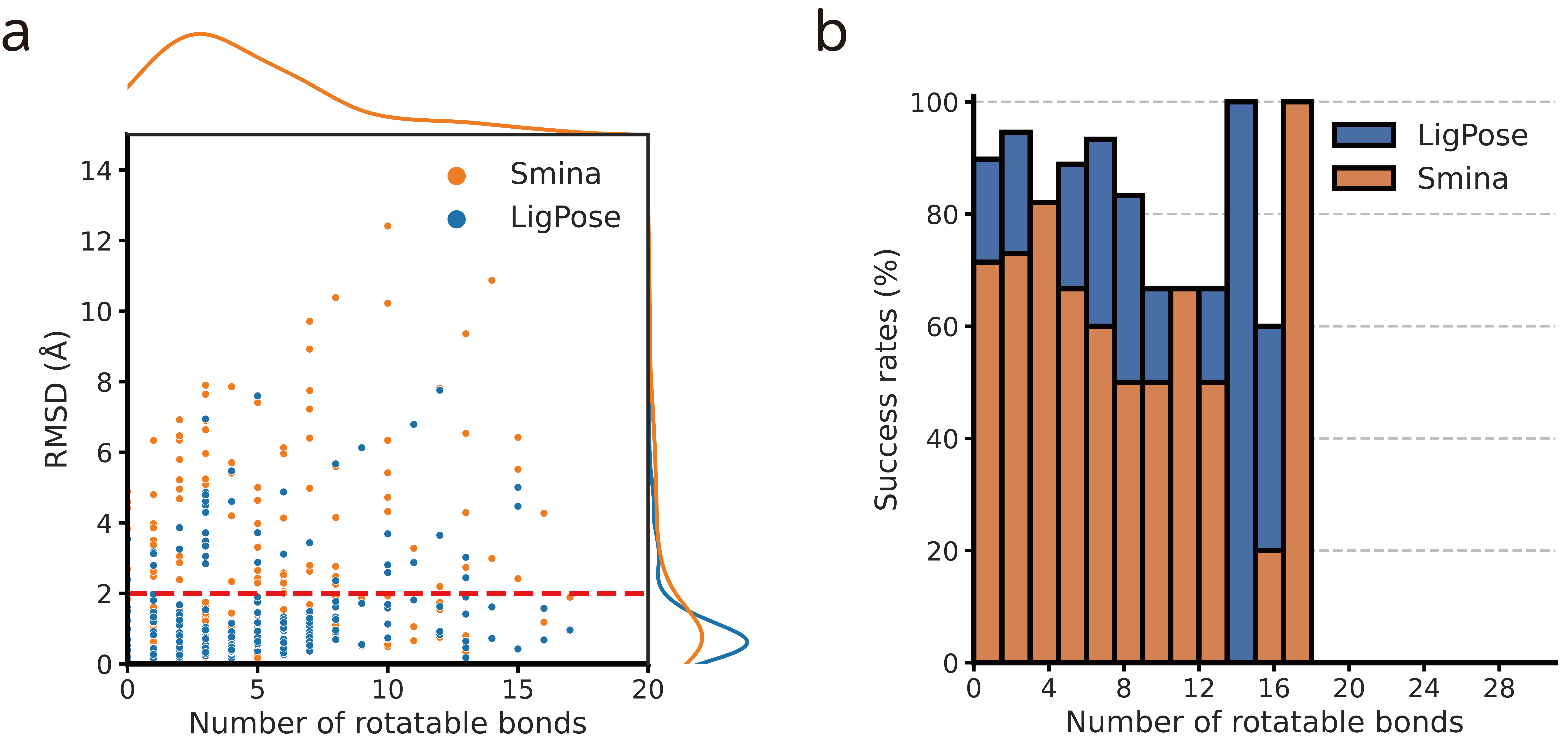}
\caption{\textbf{\textbf{(a)}, RMSD values and \textbf{(b)}, success rates for prediction of \ourmodel and Smina with respect to the number of rotatable bonds in the core set of PDBbind.} The red dashed line denotes the RMSD threshold of 2\AA. \ourmodel showed better performance in most cases. }
\label{fig:corerotbond} 
\end{figure}

\begin{figure}[h]
\centering
\includegraphics[width=0.5\linewidth]{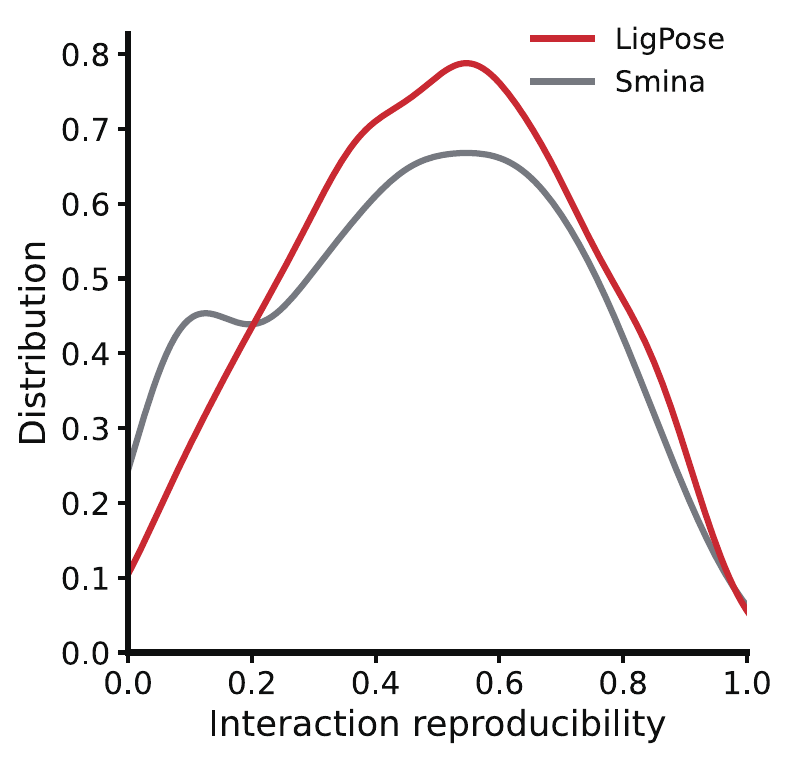}
\caption{\textbf{Distribution of the interaction reproducibility of all non-covalent interactions of \ourmodel (red line) and Smina (grey line).} The higher accuracy of \ourmodel is positively correlated to better performance on reproducing the native-like non-covalent interactions, which is important for drug screening and lead optimization. }
\label{fig:dockingnoncov1} 
\end{figure}

\begin{figure}[h]
\centering
\includegraphics[width=\linewidth]{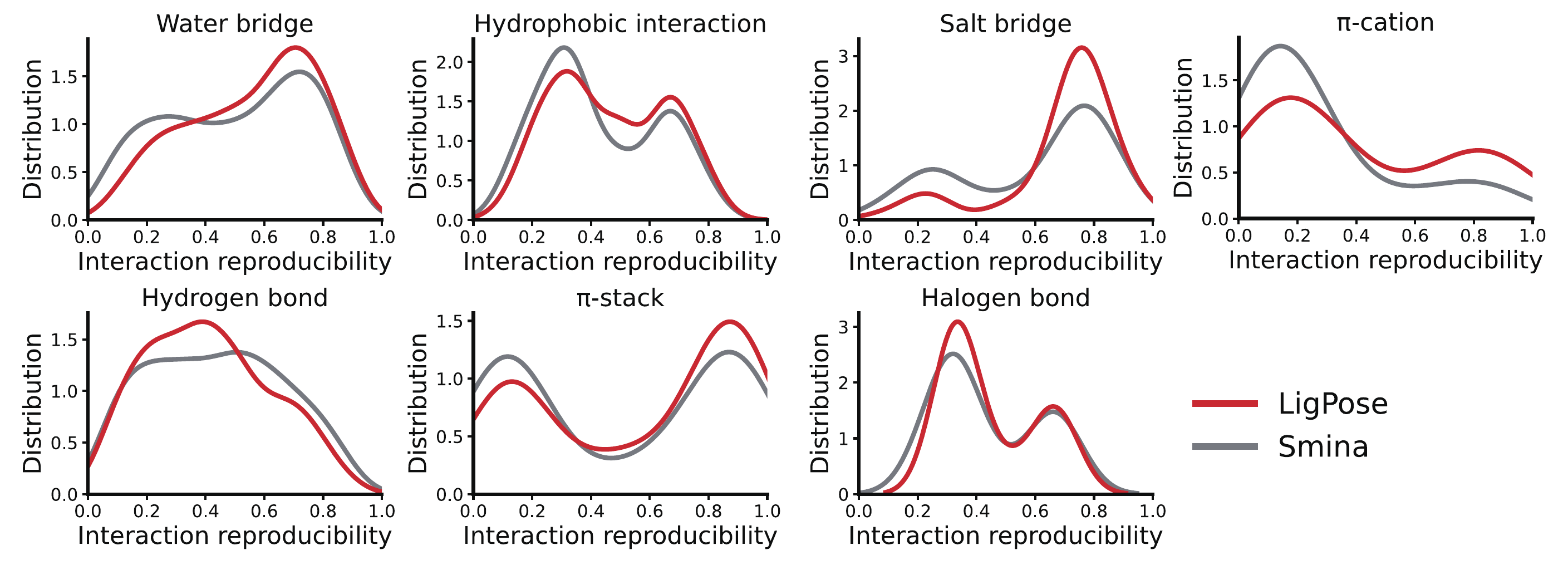}
\caption{\textbf{Distribution of the interaction reproducibility of seven non-covalent interactions of \ourmodel (red line) and Smina (grey line),} including water bridge, hydrophobic interaction, salt bridge, $\pi$-cation, hydrogen bond, $\pi$-stack and halogen bond. These results indicated that \ourmodel consistently outperformed Smina on all seven types of non-covalent interactions.}
\label{fig:dockingnoncov2} 
\end{figure}

\end{appendices}

\clearpage
\bibliographystyle{unsrt} 
\bibliography{ligpose}


\end{document}